\documentclass[10pt,twocolumn,letterpaper]{article}

\usepackage{cvpr}              

\definecolor{cvprblue}{rgb}{0.21,0.49,0.74}
\usepackage[pagebackref,breaklinks,colorlinks,allcolors=cvprblue]{hyperref}

\usepackage{url}
\usepackage{graphicx}

\usepackage[utf8]{inputenc} 
\usepackage[T1]{fontenc}    
\usepackage{booktabs}       
\usepackage{amsfonts}       
\usepackage{nicefrac}       
\usepackage{microtype}      
\usepackage{xcolor}         
\usepackage{graphicx}
\usepackage{amsmath}
\usepackage{makecell}
\usepackage{multirow}
\usepackage{enumitem}
\usepackage{adjustbox}
\usepackage{colortbl}

\def\ie{\textit{i.e.}}
\def\eg{\textit{e.g.}}
\def\vs{\textit{v.s.}}

\newcolumntype{x}[1]{>{\centering\arraybackslash}p{#1}}
\newcolumntype{y}[1]{>{\raggedright\arraybackslash}p{#1}}
\newcolumntype{z}[1]{>{\raggedleft\arraybackslash}p{#1}}
\newcommand{\tablestyle}[2]{\setlength{\tabcolsep}{#1}\renewcommand{\arraystretch}{#2}\centering\footnotesize}

\newcommand{\dataset}{EntityGrid-QA}
\newcommand{\method}{HiRes-LLaVA}
\newcommand{\adapter}{SliceRestore adapter}
\definecolor{mygray}{gray}{.9}

\renewcommand{\paragraph}[1]{\vspace{0.1mm}\noindent\textbf{#1}}

\usepackage{xspace}    
\usepackage{colortbl}
\usepackage{pifont}
\newcommand{\cmark}{\ding{51}\xspace}%
\newcommand{\xmark}{\ding{55}\xspace}%
\newcommand{\xmarkg}{\textcolor{lightgray}{\ding{55}}\xspace}%
\definecolor{mygreen}{rgb}{0.0, 0.51, 0.0}
\newcommand{\improve}[1]{\bf \textcolor{mygreen}{#1}}
\definecolor{citecolor}{HTML}{0071bc}

\def\paperID{1824} 
\def\confName{CVPR}
\def\confYear{2025}

\title{HiRes-LLaVA: Restoring Fragmentation Input \\ in
High-Resolution Large Vision-Language Models}

\author{%
Runhui Huang$^{1}$\thanks{Equal contribution} ~ \ Xinpeng Ding$^{3*}$ \ Chunwei Wang$^{2}$ \ Jianhua Han$^{2}$ \\
Yulong Liu$^{3}$ \ Hengshuang Zhao$^{4}$ \ Hang Xu$^{2}$ \ Lu Hou$^{2}$ \ Wei Zhang$^{2}$ \ Xiaodan Liang$^{1}$\footnotemark[2] \\
\textsuperscript{1}Shenzhen campus of Sun Yat-sen University \ \textsuperscript{2}Huawei \\
\textsuperscript{3}The Hong Kong University of Science and Technology \ 
\textsuperscript{4}The University of Hong Kong 
}

\begin{document}
\maketitle


\newcommand{\figleft}{{\em (Left)}}
\newcommand{\figcenter}{{\em (Center)}}
\newcommand{\figright}{{\em (Right)}}
\newcommand{\figtop}{{\em (Top)}}
\newcommand{\figbottom}{{\em (Bottom)}}
\newcommand{\captiona}{{\em (a)}}
\newcommand{\captionb}{{\em (b)}}
\newcommand{\captionc}{{\em (c)}}
\newcommand{\captiond}{{\em (d)}}

\newcommand{\newterm}[1]{{\bf #1}}

\def\figref#1{figure~\ref{#1}}
\def\Figref#1{Figure~\ref{#1}}
\def\twofigref#1#2{figures \ref{#1} and \ref{#2}}
\def\quadfigref#1#2#3#4{figures \ref{#1}, \ref{#2}, \ref{#3} and \ref{#4}}
\def\secref#1{section~\ref{#1}}
\def\Secref#1{Section~\ref{#1}}
\def\twosecrefs#1#2{sections \ref{#1} and \ref{#2}}
\def\secrefs#1#2#3{sections \ref{#1}, \ref{#2} and \ref{#3}}
\def\eqref#1{equation~\ref{#1}}
\def\Eqref#1{Equation~\ref{#1}}
\def\plaineqref#1{\ref{#1}}
\def\chapref#1{chapter~\ref{#1}}
\def\Chapref#1{Chapter~\ref{#1}}
\def\rangechapref#1#2{chapters\ref{#1}--\ref{#2}}
\def\algref#1{algorithm~\ref{#1}}
\def\Algref#1{Algorithm~\ref{#1}}
\def\twoalgref#1#2{algorithms \ref{#1} and \ref{#2}}
\def\Twoalgref#1#2{Algorithms \ref{#1} and \ref{#2}}
\def\partref#1{part~\ref{#1}}
\def\Partref#1{Part~\ref{#1}}
\def\twopartref#1#2{parts \ref{#1} and \ref{#2}}

\def\ceil#1{\lceil #1 \rceil}
\def\floor#1{\lfloor #1 \rfloor}
\def\1{\bm{1}}
\newcommand{\train}{\mathcal{D}}
\newcommand{\valid}{\mathcal{D_{\mathrm{valid}}}}
\newcommand{\test}{\mathcal{D_{\mathrm{test}}}}

\def\eps{{\epsilon}}

\def\reta{{\textnormal{$\eta$}}}
\def\ra{{\textnormal{a}}}
\def\rb{{\textnormal{b}}}
\def\rc{{\textnormal{c}}}
\def\rd{{\textnormal{d}}}
\def\re{{\textnormal{e}}}
\def\rf{{\textnormal{f}}}
\def\rg{{\textnormal{g}}}
\def\rh{{\textnormal{h}}}
\def\ri{{\textnormal{i}}}
\def\rj{{\textnormal{j}}}
\def\rk{{\textnormal{k}}}
\def\rl{{\textnormal{l}}}
\def\rn{{\textnormal{n}}}
\def\ro{{\textnormal{o}}}
\def\rp{{\textnormal{p}}}
\def\rq{{\textnormal{q}}}
\def\rr{{\textnormal{r}}}
\def\rs{{\textnormal{s}}}
\def\rt{{\textnormal{t}}}
\def\ru{{\textnormal{u}}}
\def\rv{{\textnormal{v}}}
\def\rw{{\textnormal{w}}}
\def\rx{{\textnormal{x}}}
\def\ry{{\textnormal{y}}}
\def\rz{{\textnormal{z}}}

\def\rvepsilon{{\mathbf{\epsilon}}}
\def\rvtheta{{\mathbf{\theta}}}
\def\rva{{\mathbf{a}}}
\def\rvb{{\mathbf{b}}}
\def\rvc{{\mathbf{c}}}
\def\rvd{{\mathbf{d}}}
\def\rve{{\mathbf{e}}}
\def\rvf{{\mathbf{f}}}
\def\rvg{{\mathbf{g}}}
\def\rvh{{\mathbf{h}}}
\def\rvu{{\mathbf{i}}}
\def\rvj{{\mathbf{j}}}
\def\rvk{{\mathbf{k}}}
\def\rvl{{\mathbf{l}}}
\def\rvm{{\mathbf{m}}}
\def\rvn{{\mathbf{n}}}
\def\rvo{{\mathbf{o}}}
\def\rvp{{\mathbf{p}}}
\def\rvq{{\mathbf{q}}}
\def\rvr{{\mathbf{r}}}
\def\rvs{{\mathbf{s}}}
\def\rvt{{\mathbf{t}}}
\def\rvu{{\mathbf{u}}}
\def\rvv{{\mathbf{v}}}
\def\rvw{{\mathbf{w}}}
\def\rvx{{\mathbf{x}}}
\def\rvy{{\mathbf{y}}}
\def\rvz{{\mathbf{z}}}

\def\erva{{\textnormal{a}}}
\def\ervb{{\textnormal{b}}}
\def\ervc{{\textnormal{c}}}
\def\ervd{{\textnormal{d}}}
\def\erve{{\textnormal{e}}}
\def\ervf{{\textnormal{f}}}
\def\ervg{{\textnormal{g}}}
\def\ervh{{\textnormal{h}}}
\def\ervi{{\textnormal{i}}}
\def\ervj{{\textnormal{j}}}
\def\ervk{{\textnormal{k}}}
\def\ervl{{\textnormal{l}}}
\def\ervm{{\textnormal{m}}}
\def\ervn{{\textnormal{n}}}
\def\ervo{{\textnormal{o}}}
\def\ervp{{\textnormal{p}}}
\def\ervq{{\textnormal{q}}}
\def\ervr{{\textnormal{r}}}
\def\ervs{{\textnormal{s}}}
\def\ervt{{\textnormal{t}}}
\def\ervu{{\textnormal{u}}}
\def\ervv{{\textnormal{v}}}
\def\ervw{{\textnormal{w}}}
\def\ervx{{\textnormal{x}}}
\def\ervy{{\textnormal{y}}}
\def\ervz{{\textnormal{z}}}

\def\rmA{{\mathbf{A}}}
\def\rmB{{\mathbf{B}}}
\def\rmC{{\mathbf{C}}}
\def\rmD{{\mathbf{D}}}
\def\rmE{{\mathbf{E}}}
\def\rmF{{\mathbf{F}}}
\def\rmG{{\mathbf{G}}}
\def\rmH{{\mathbf{H}}}
\def\rmI{{\mathbf{I}}}
\def\rmJ{{\mathbf{J}}}
\def\rmK{{\mathbf{K}}}
\def\rmL{{\mathbf{L}}}
\def\rmM{{\mathbf{M}}}
\def\rmN{{\mathbf{N}}}
\def\rmO{{\mathbf{O}}}
\def\rmP{{\mathbf{P}}}
\def\rmQ{{\mathbf{Q}}}
\def\rmR{{\mathbf{R}}}
\def\rmS{{\mathbf{S}}}
\def\rmT{{\mathbf{T}}}
\def\rmU{{\mathbf{U}}}
\def\rmV{{\mathbf{V}}}
\def\rmW{{\mathbf{W}}}
\def\rmX{{\mathbf{X}}}
\def\rmY{{\mathbf{Y}}}
\def\rmZ{{\mathbf{Z}}}

\def\ermA{{\textnormal{A}}}
\def\ermB{{\textnormal{B}}}
\def\ermC{{\textnormal{C}}}
\def\ermD{{\textnormal{D}}}
\def\ermE{{\textnormal{E}}}
\def\ermF{{\textnormal{F}}}
\def\ermG{{\textnormal{G}}}
\def\ermH{{\textnormal{H}}}
\def\ermI{{\textnormal{I}}}
\def\ermJ{{\textnormal{J}}}
\def\ermK{{\textnormal{K}}}
\def\ermL{{\textnormal{L}}}
\def\ermM{{\textnormal{M}}}
\def\ermN{{\textnormal{N}}}
\def\ermO{{\textnormal{O}}}
\def\ermP{{\textnormal{P}}}
\def\ermQ{{\textnormal{Q}}}
\def\ermR{{\textnormal{R}}}
\def\ermS{{\textnormal{S}}}
\def\ermT{{\textnormal{T}}}
\def\ermU{{\textnormal{U}}}
\def\ermV{{\textnormal{V}}}
\def\ermW{{\textnormal{W}}}
\def\ermX{{\textnormal{X}}}
\def\ermY{{\textnormal{Y}}}
\def\ermZ{{\textnormal{Z}}}

\def\vzero{{\bm{0}}}
\def\vone{{\bm{1}}}
\def\vmu{{\bm{\mu}}}
\def\vtheta{{\bm{\theta}}}
\def\va{{\bm{a}}}
\def\vb{{\bm{b}}}
\def\vc{{\bm{c}}}
\def\vd{{\bm{d}}}
\def\ve{{\bm{e}}}
\def\vf{{\bm{f}}}
\def\vg{{\bm{g}}}
\def\vh{{\bm{h}}}
\def\vi{{\bm{i}}}
\def\vj{{\bm{j}}}
\def\vk{{\bm{k}}}
\def\vl{{\bm{l}}}
\def\vm{{\bm{m}}}
\def\vn{{\bm{n}}}
\def\vo{{\bm{o}}}
\def\vp{{\bm{p}}}
\def\vq{{\bm{q}}}
\def\vr{{\bm{r}}}
\def\vs{{\bm{s}}}
\def\vt{{\bm{t}}}
\def\vu{{\bm{u}}}
\def\vv{{\bm{v}}}
\def\vw{{\bm{w}}}
\def\vx{{\bm{x}}}
\def\vy{{\bm{y}}}
\def\vz{{\bm{z}}}

\def\evalpha{{\alpha}}
\def\evbeta{{\beta}}
\def\evepsilon{{\epsilon}}
\def\evlambda{{\lambda}}
\def\evomega{{\omega}}
\def\evmu{{\mu}}
\def\evpsi{{\psi}}
\def\evsigma{{\sigma}}
\def\evtheta{{\theta}}
\def\eva{{a}}
\def\evb{{b}}
\def\evc{{c}}
\def\evd{{d}}
\def\eve{{e}}
\def\evf{{f}}
\def\evg{{g}}
\def\evh{{h}}
\def\evi{{i}}
\def\evj{{j}}
\def\evk{{k}}
\def\evl{{l}}
\def\evm{{m}}
\def\evn{{n}}
\def\evo{{o}}
\def\evp{{p}}
\def\evq{{q}}
\def\evr{{r}}
\def\evs{{s}}
\def\evt{{t}}
\def\evu{{u}}
\def\evv{{v}}
\def\evw{{w}}
\def\evx{{x}}
\def\evy{{y}}
\def\evz{{z}}

\def\mA{{\bm{A}}}
\def\mB{{\bm{B}}}
\def\mC{{\bm{C}}}
\def\mD{{\bm{D}}}
\def\mE{{\bm{E}}}
\def\mF{{\bm{F}}}
\def\mG{{\bm{G}}}
\def\mH{{\bm{H}}}
\def\mI{{\bm{I}}}
\def\mJ{{\bm{J}}}
\def\mK{{\bm{K}}}
\def\mL{{\bm{L}}}
\def\mM{{\bm{M}}}
\def\mN{{\bm{N}}}
\def\mO{{\bm{O}}}
\def\mP{{\bm{P}}}
\def\mQ{{\bm{Q}}}
\def\mR{{\bm{R}}}
\def\mS{{\bm{S}}}
\def\mT{{\bm{T}}}
\def\mU{{\bm{U}}}
\def\mV{{\bm{V}}}
\def\mW{{\bm{W}}}
\def\mX{{\bm{X}}}
\def\mY{{\bm{Y}}}
\def\mZ{{\bm{Z}}}
\def\mBeta{{\bm{\beta}}}
\def\mPhi{{\bm{\Phi}}}
\def\mLambda{{\bm{\Lambda}}}
\def\mSigma{{\bm{\Sigma}}}

\newcommand{\tens}[1]{\bm{\mathsfit{#1}}}
\def\tA{{\tens{A}}}
\def\tB{{\tens{B}}}
\def\tC{{\tens{C}}}
\def\tD{{\tens{D}}}
\def\tE{{\tens{E}}}
\def\tF{{\tens{F}}}
\def\tG{{\tens{G}}}
\def\tH{{\tens{H}}}
\def\tI{{\tens{I}}}
\def\tJ{{\tens{J}}}
\def\tK{{\tens{K}}}
\def\tL{{\tens{L}}}
\def\tM{{\tens{M}}}
\def\tN{{\tens{N}}}
\def\tO{{\tens{O}}}
\def\tP{{\tens{P}}}
\def\tQ{{\tens{Q}}}
\def\tR{{\tens{R}}}
\def\tS{{\tens{S}}}
\def\tT{{\tens{T}}}
\def\tU{{\tens{U}}}
\def\tV{{\tens{V}}}
\def\tW{{\tens{W}}}
\def\tX{{\tens{X}}}
\def\tY{{\tens{Y}}}
\def\tZ{{\tens{Z}}}

\def\gA{{\mathcal{A}}}
\def\gB{{\mathcal{B}}}
\def\gC{{\mathcal{C}}}
\def\gD{{\mathcal{D}}}
\def\gE{{\mathcal{E}}}
\def\gF{{\mathcal{F}}}
\def\gG{{\mathcal{G}}}
\def\gH{{\mathcal{H}}}
\def\gI{{\mathcal{I}}}
\def\gJ{{\mathcal{J}}}
\def\gK{{\mathcal{K}}}
\def\gL{{\mathcal{L}}}
\def\gM{{\mathcal{M}}}
\def\gN{{\mathcal{N}}}
\def\gO{{\mathcal{O}}}
\def\gP{{\mathcal{P}}}
\def\gQ{{\mathcal{Q}}}
\def\gR{{\mathcal{R}}}
\def\gS{{\mathcal{S}}}
\def\gT{{\mathcal{T}}}
\def\gU{{\mathcal{U}}}
\def\gV{{\mathcal{V}}}
\def\gW{{\mathcal{W}}}
\def\gX{{\mathcal{X}}}
\def\gY{{\mathcal{Y}}}
\def\gZ{{\mathcal{Z}}}

\def\sA{{\mathbb{A}}}
\def\sB{{\mathbb{B}}}
\def\sC{{\mathbb{C}}}
\def\sD{{\mathbb{D}}}
\def\sF{{\mathbb{F}}}
\def\sG{{\mathbb{G}}}
\def\sH{{\mathbb{H}}}
\def\sI{{\mathbb{I}}}
\def\sJ{{\mathbb{J}}}
\def\sK{{\mathbb{K}}}
\def\sL{{\mathbb{L}}}
\def\sM{{\mathbb{M}}}
\def\sN{{\mathbb{N}}}
\def\sO{{\mathbb{O}}}
\def\sP{{\mathbb{P}}}
\def\sQ{{\mathbb{Q}}}
\def\sR{{\mathbb{R}}}
\def\sS{{\mathbb{S}}}
\def\sT{{\mathbb{T}}}
\def\sU{{\mathbb{U}}}
\def\sV{{\mathbb{V}}}
\def\sW{{\mathbb{W}}}
\def\sX{{\mathbb{X}}}
\def\sY{{\mathbb{Y}}}
\def\sZ{{\mathbb{Z}}}

\def\emLambda{{\Lambda}}
\def\emA{{A}}
\def\emB{{B}}
\def\emC{{C}}
\def\emD{{D}}
\def\emE{{E}}
\def\emF{{F}}
\def\emG{{G}}
\def\emH{{H}}
\def\emI{{I}}
\def\emJ{{J}}
\def\emK{{K}}
\def\emL{{L}}
\def\emM{{M}}
\def\emN{{N}}
\def\emO{{O}}
\def\emP{{P}}
\def\emQ{{Q}}
\def\emR{{R}}
\def\emS{{S}}
\def\emT{{T}}
\def\emU{{U}}
\def\emV{{V}}
\def\emW{{W}}
\def\emX{{X}}
\def\emY{{Y}}
\def\emZ{{Z}}
\def\emSigma{{\Sigma}}

\newcommand{\etens}[1]{\mathsfit{#1}}
\def\etLambda{{\etens{\Lambda}}}
\def\etA{{\etens{A}}}
\def\etB{{\etens{B}}}
\def\etC{{\etens{C}}}
\def\etD{{\etens{D}}}
\def\etE{{\etens{E}}}
\def\etF{{\etens{F}}}
\def\etG{{\etens{G}}}
\def\etH{{\etens{H}}}
\def\etI{{\etens{I}}}
\def\etJ{{\etens{J}}}
\def\etK{{\etens{K}}}
\def\etL{{\etens{L}}}
\def\etM{{\etens{M}}}
\def\etN{{\etens{N}}}
\def\etO{{\etens{O}}}
\def\etP{{\etens{P}}}
\def\etQ{{\etens{Q}}}
\def\etR{{\etens{R}}}
\def\etS{{\etens{S}}}
\def\etT{{\etens{T}}}
\def\etU{{\etens{U}}}
\def\etV{{\etens{V}}}
\def\etW{{\etens{W}}}
\def\etX{{\etens{X}}}
\def\etY{{\etens{Y}}}
\def\etZ{{\etens{Z}}}

\newcommand{\pdata}{p_{\rm{data}}}
\newcommand{\ptrain}{\hat{p}_{\rm{data}}}
\newcommand{\Ptrain}{\hat{P}_{\rm{data}}}
\newcommand{\pmodel}{p_{\rm{model}}}
\newcommand{\Pmodel}{P_{\rm{model}}}
\newcommand{\ptildemodel}{\tilde{p}_{\rm{model}}}
\newcommand{\pencode}{p_{\rm{encoder}}}
\newcommand{\pdecode}{p_{\rm{decoder}}}
\newcommand{\precons}{p_{\rm{reconstruct}}}

\newcommand{\laplace}{\mathrm{Laplace}} 

\newcommand{\E}{\mathbb{E}}
\newcommand{\Ls}{\mathcal{L}}
\newcommand{\R}{\mathbb{R}}
\newcommand{\emp}{\tilde{p}}
\newcommand{\lr}{\alpha}
\newcommand{\reg}{\lambda}
\newcommand{\rect}{\mathrm{rectifier}}
\newcommand{\softmax}{\mathrm{softmax}}
\newcommand{\sigmoid}{\sigma}
\newcommand{\softplus}{\zeta}
\newcommand{\KL}{D_{\mathrm{KL}}}
\newcommand{\Var}{\mathrm{Var}}
\newcommand{\standarderror}{\mathrm{SE}}
\newcommand{\Cov}{\mathrm{Cov}}
\newcommand{\normlzero}{L^0}
\newcommand{\normlone}{L^1}
\newcommand{\normltwo}{L^2}
\newcommand{\normlp}{L^p}
\newcommand{\normmax}{L^\infty}

\newcommand{\parents}{Pa} 

\let\ab\allowbreak

\begin{abstract}
High-resolution image inputs allow Large Vision-Language Models (LVLMs) to capture finer visual details, improving comprehension. However, the increased training and computational costs associated with such inputs pose significant challenges. A common approach to mitigate these costs involves slicing the input into uniform patches using sliding windows, each aligned with the vision encoder’s input size. While efficient, this method fragments the input, disrupting the continuity of context, which negatively impacts cross-patch perception tasks.
To address these limitations, we propose \textbf{\method}, a novel framework designed to efficiently process high-resolution inputs of any size without altering the original contextual and geometric information. \method~introduces two key components: (i) a SliceRestore Adapter (SRA) that reconstructs sliced patches into their original form, enabling efficient extraction of both global and local features through down-up-sampling and convolutional layers, and (ii) a Self-Mining Sampler (SMS) that compresses visual tokens based on internal relationships, preserving original context and positional information while reducing training overhead.
To assess the ability of handling context fragmentation, we construct a new benchmark, \dataset, consisting of edge-related tasks. 
Extensive experiments demonstrate the superiority of \method~on both existing public benchmarks and \dataset. For example, with SRA, our method achieves a performance improvement of $\sim12\%$ over state-of-the-art LVLMs in addressing fragmentation issues. Additionally, our SMS outperforms other visual token downsamplers, while offering high data efficiency.
\end{abstract}
\section{Introduction}
Recent progress in Large Vision-Language Models (LVLMs)~\citep{alayrac2022flamingo,li2023blip,li2023llava,li2023videochat,liu2023visual,zhu2023minigpt} has significantly enhanced capabilities in vision-language tasks, fostering improved understanding, reasoning, and interaction.
Early LVLMs~\citep{li2023llava,zhu2023minigpt,liu2023improved} processed images at low resolutions, typically $224 \times 224$, which hindering their ability to capture detailed visual information. This limitation often results in inaccurate recognition of objects and their contextual relationships within images~\citep{ding2023hilm,li2023monkey}.

\begin{figure}[t]
    \centering
\includegraphics[width=\linewidth,height=0.2\textheight]{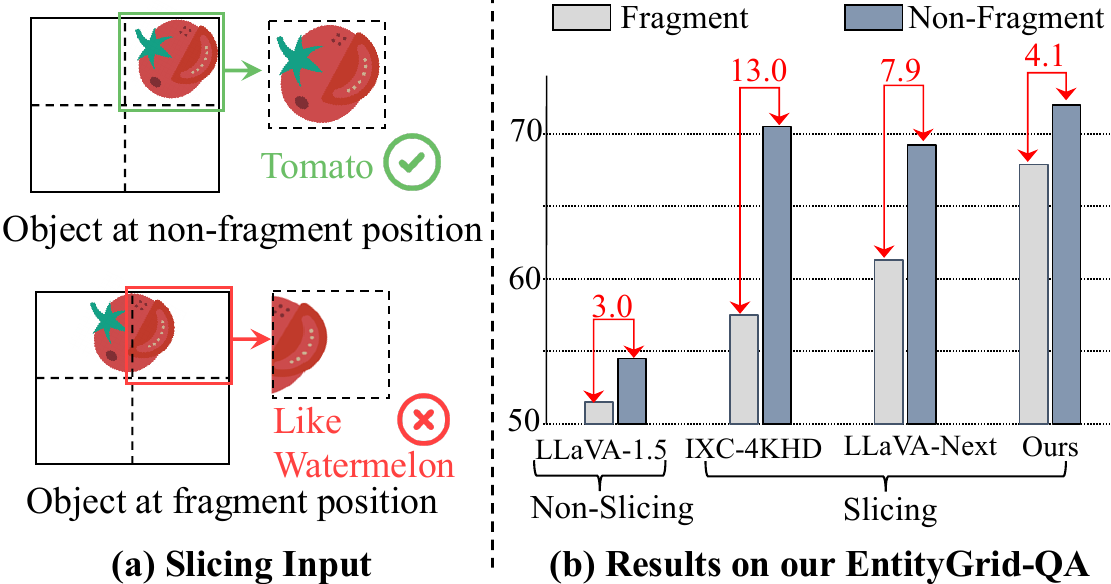}
   \caption{\textbf{Illustration of the fragmentation issue.} \textbf{(a) Slicing input:} Slicing-based LVLMs, such as LLaVA-Next~\cite{liu2024llavanext}, can fragment objects located at the edges of slices, leading to errors in model understanding. \textbf{(b) Performance comparison:} On our EntityGrid-QA benchmark, slicing-based methods show a significant performance gap between fragment and non-fragment inputs. Our method effectively handles both cases, achieving a smaller performance gap similar to non-slicing approaches.}
    \label{fig:introduction}
    \vspace{-2mm}
\end{figure}

Enhancing the high-resolution capabilities of LVLMs presents substantial challenges,~\ie, training visual encoders to handle high-resolution inputs requires significant computational resources as well as struggling with handling arbitrary image sizes~\citep{Qwen-VL,chen2023pali}.
Recent advances have introduced resource-efficient methods to improve the input resolution of LVLMs. One effective strategy involves using a sliding window technique~\citep{li2023monkey,xu2024llava,liu2024textmonkey} to segment high-resolution images into smaller patches. These patches are then processed by a visual encoder that has been trained on fixed-size lower-resolution inputs, maintaining computational efficiency while enhancing detail capture.

Slicing-based approaches, while effective, can lead to input fragmentation, as shown in \cref{fig:introduction}~(a). When objects are located at the edges of slices, the disrupted context can cause misclassifications, such as labeling a `Tomato' as a `Watermelon'. This fragmentation compromises spatial relationships and semantic coherence, challenging the model's understanding.
To validate this, we compare non-slicing-based LVLMs, such as LLaVA-1.5~\cite{liu2023improved}, with slicing-based methods like IXC-4KHD~\cite{zhang2023internlm} and LLaVA-Next~\cite{liu2024llavanext} under fragmented and non-fragmented scenarios. As shown in \cref{fig:introduction}~(b), slicing-based methods exhibit a large performance gap ($13.0\%$), while non-slicing-based methods are more consistent (only $3.0\%$ gap), underscoring the limitations of slicing in preserving context integrity.
Furthermore, existing methods~\cite{xu2024llava,liu2024textmonkey} also rely on Q-Former-like samplers~\cite{li2023blip} to handle long contexts from high-resolution inputs. However, these samplers suffer from severe drawbacks, such as the lack of positional information and high training overhead~\cite{yao2024deco}, making them suboptimal for context-rich scenarios.

In this paper, we propose \method, an efficient approach to integrating high-resolution data into LVLMs without disrupting the original context. 
\method~introduces the SliceRestore Adapter for the vision encoder to allow the high-resolution slices to preserve the image's complete context and capture the edge information among slices. This enhancement is processed through dual fusion modules to capture both global and local information in parallel.
This lightweight module can be seamlessly integrated into any attention layer of the vision encoder for extracting the features of high-resolution images, enabling efficient fine-tuning without altering pre-trained parameters. 
Furthermore, \method~introduces the self-mining sampler that uses pooled sliced patches as queries to compress visual tokens from non-overlapped areas. Unlike fixed learnable query-based methods, our self-mining sampler not only preserves the original context and positional information but also performs high data efficiency.

To evaluate our proposed method, we tested it on nine widely-used public benchmarks and also introduced a new benchmark, \dataset, specifically designed to measure how well VLMs handle context fragmentation caused by slicing approaches. Our comprehensive experiments show that \method~not only performs better than current models on these public benchmarks but also significantly surpasses SOTA LVLMs over $12\%$ on the \dataset. Additionally, our SMS outperforms other visual token downsampling methods and improves 40\% data efficiency.

\section{Related Works}
\noindent\textbf{Large vision-language model.}
Leveraging pre-trained Large Language Models (LLMs) like LLaMA~\citep{touvron2023llama} and Vicuna~\citep{chiang2023vicuna}, LVLMs have achieved significant advancements in areas such as image/video understanding~\citep{li2022blip,li2023blip,zhu2023minigpt,alayrac2022flamingo,chen2023shikra,zhang2023video,li2023videochat}, medical analysis~\citep{li2023llava}, and autonomous driving~\citep{ding2023hilm,xu2023drivegpt4}.
These models use vision encoders trained via contrastive learning~\citep{dosovitskiy2020image,radford2021learning} to align visual features with language. Visual embeddings are adapted to match the LLM dimensionality using visual projectors. These projectors can be: (i) resamplers, like Q-Former~\citep{alayrac2022flamingo,li2023blip,zhu2023minigpt}, using fixed queries for cross-attention, or (ii) MLP modules, as seen in the LLaVA series~\citep{liu2023visual}.
Recent efforts have aimed to enhance visual representation by combining features from DINO-V2~\citep{oquab2023dinov2} and SAM~\citep{kirillov2023segany} with CLIP’s Vision Transformers (ViT)~\citep{Ranzinger_Heinrich_Kautz_Molchanov_2023,Lin_Liu_Zhang_Gao_Qiu_Xiao_Qiu_Lin_Shao_Chen_et}. However, CLIP-ViT's fixed-resolution requirement (\eg, $336 \times 336$) limits the capability to handle higher resolution and varying aspect ratios, thereby hindering performance in fine-grained tasks.

\noindent\textbf{High-resolution large vision-language model.}
To discern fine-grained visual details from high-resolution inputs, an intuitive approach is to split images into patches and project them using linear layers, treating these as a sequence for input into Large Vision-Language Models (LVLMs)~\citep{fuyu-8b,li2023otterhd}. While this eliminates the need for an image encoder, it often results in insufficient visual representation, leading to increased training costs and suboptimal performance.
Alternatively, Up-Resize methods such as Qwen-VL~\citep{Qwen-VL} adapt the positional embeddings of ViT from $224 \times 224$ to $448 \times 448$ and include an additional training phase to fine-tune the ViT. However, this adaptation may alter the original visual position encoding from CLIP-ViT~\citep{radford2021learning}, potentially degrading visual representation.
Dual-branch approaches introduce a high-resolution branch with lightweight convolutional networks to manage high-resolution inputs but require additional training data and parameters~\citep{hong2023cogagent,ding2023hilm,luo2024feast,li2024mini}.
Slicing-based methods offer a compromise by using slicing windows to divide the high-resolution image into patches that match the input size of a pre-trained vision encoder, maintaining efficiency in parameter use and training data while still achieving competitive performance~\citep{li2023monkey,xu2024llava}. However, they suffer from "Context Fragmentation", where the continuity of contextual information across patches is damaged, impacting tasks that require cross-patch context.
In this paper, we propose \method, a novel technique designed to seamlessly integrate global-local high-resolution details into LVLMs without disrupting the original context, effectively addressing the issue of Context Fragmentation.

\section{Method}
In this section, we first present the overall framework of \method~ in \cref{sec:framework}. The two innovative components, namely \adapter~ and self-mining sampler are detailed in \cref{sec:adapter} and \cref{sec:sampler} respectively. To further evaluate the ability of VLMs to address the context fragmentation issue, a new benchmark named EntityGrid-QA is proposed in \cref{sec:benchmark}.

\begin{figure*}[t!]
\begin{center}
\includegraphics[width=1\textwidth]{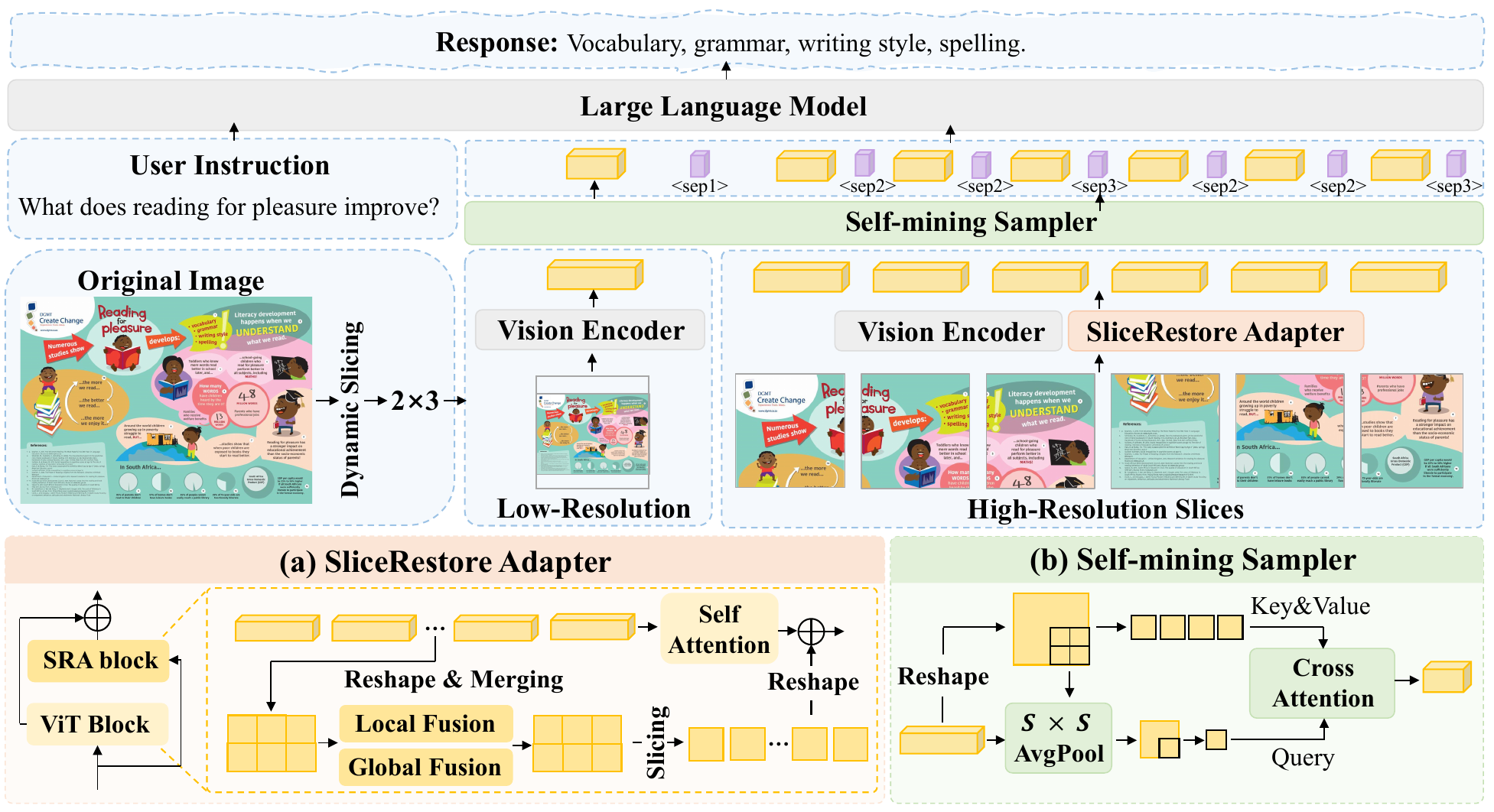}
\caption{
\textbf{Overall framework of \method.} The vision encoding consists of two branches: one for low-resolution images processed by the pre-trained vision encoder to extract global features, and another dividing high-resolution images into multiple slices to capture fine-grained details.
\textbf{(a) SliceRestore Adapter} aims to address the Context Fragmentation issue, it restores sliced features into a whole feature by capturing both local and global information, then splits the whole feature back into slices.
\textbf{(b) Self-Mining Sampler} compresses visual token numbers to reduce computation and memory costs by using downsampled features as queries and the original features as keys and values. Both low-resolution image input and each high-resolution slice are compressed by the same self-mining sampler.
}
\label{fig:framework}
\vspace{-6mm}
\end{center}
\end{figure*}
\subsection{Overall Framework}~\label{sec:framework}
The overall framework of \method~is shown in \cref{fig:framework}. First, the original image is resized and padded to a low resolution (typically $224 \times 224$), then processed by the pre-trained vision encoder to produce global features. To capture fine-grained details, the high-resolution image is split into smaller slices. 

Specifically, we set a maximum slice count $M$, allowing each image to automatically select an optimal slicing grid with $m$ columns and $n$ rows slices. The values of $m$ and $n$ are determined based on the base resolution $r$ of the pretrained vision encoder as follows:
\begin{equation}
    m = \left\lceil \frac{H}{r} \right\rceil, \quad n = \left\lceil \frac{W}{r} \right\rceil,
    \label{eq:mn_def}
\end{equation}
This slicing approach adapts to the original aspect ratio of the image. If the resulting number of slices ($4 \times m \times n$) does not exceed the maximum slice count $M$, the $m$ and $n$ are scale up by a factor of 2, ensuring detailed preservation without overwhelming the model.

After slicing, these slices are processed by a shared vision encoder with the proposed \adapter, yielding slice features, followed by a shared self-mining sampler to reduce token length, resulting in compressed features. Consequently, the visual input to the language model includes a low-resolution overview and multiple high-resolution slices. To maintain clarity, three types of separators are used to maintain clarity in (1) between the low-resolution image and high-resolution slices, (2) between high resolutions slices and (3) the end of each slice row. 

\subsection{SliceRestore Adapter}~\label{sec:adapter}
We denote the slice features in the $l$-th layer of ViT as 
$\{ \mathbf{P}_i \}_{i=1}^N$ with $\mathbf{P}_i \in \mathbb{R}^{L \times D}$, where $N$ is the number of slices, $L = H \times W$ is the token length, and $D$ is the feature dimension. Each slice feature is processed individually by the self-attention layer, $\text{\it Self-Attn}(\mathbf{P}_i)$, which lead to a loss of global information in fragmented context.
(see \cref{fig:introduction}~(a)). Although low-resolution inputs contain the overall information, when it comes to real-world scenes, the high-resolution inputs are still needed to perceive the small objects. A naive approach would be concatenating slice features for self-attention, but this incurs quadratic computation costs.

In this paper, we propose the SliceRestore Adapter~(SRA) to efficiently capture complete information from high-resolution inputs. As depicted in \cref{fig:framework}~(a), the \adapter~is integrated into the self-attention layer of vision transformer. This can be formulated as: 
\begin{equation}
 \{ \hat{\mathbf{P}}_i \}_{i=1}^{N} = \{ \mathbf{P}_i \}_{i=1}^{N} + \{ \overline{\mathbf{P}}^l_i \}_{i=1}^N,
\end{equation}
where:
\begin{equation}
 \{ \overline{\mathbf{P}}^l_i \}_{i=1}^N = \text{\it SRA}(\{ \mathbf{P}_i \}_{i=1}^N),
\end{equation}

The \adapter~has three main steps to restore complete semantics from slice features:

\noindent~\textbf{1. Merging}: Each slice feature $\mathbf{P}_i$ is reshaped to $r \times r \times D$. These reshaped slice features are then merged to recover the original spatial structure, forming the input’s features $\mathbf{F} \in \mathbb{R}^{(m*r) \times (n*r) \times D}$.

\noindent~\textbf{2. Capturing}: We propose two fusion modules that operate in parallel to capture both local and global context from $\mathbf{F}$. 
The local fusion module transfers edge details among slices to facilitate a nuanced exchange of local information. On the other hand, the global fusion module is leveraged to capture broader contextual cues. To achieve this, 
The local fusion module uses a single layer depth-wise convolution with $3\times3$ kernel and stride of 1 to efficiently capture local details and retain image-related biases. Due to the high computation cost of self-attention on high-resolution image, the global fusion module employs self-attention on the coarse view of the high-resolution image to transfer the global context to slices. The coarse view image with the same resolution of the low-resolution image, can be simply obtained by downsampling $\mathbf{F}^l$. After the attention block, the fused global feature is upsampled back to the original size using simple interpolation. The enhanced feature $\overline{\mathbf{F}}$ is obtained by element-wise addition of the outputs from the local and global fusion modules:
\begin{equation}
       \overline{\mathbf{F}}  = \underbrace{\text{\it DWConv}(\mathbf{F})}_{\text{local fusion}}  +  \underbrace{\text{\it Up}(\text{\it Self-Attn}(\text{\it Down}(\mathbf{F})))}_{\text{global fusion}}.
       \label{e:fusion}
\end{equation}

\noindent~\textbf{3. Slicing}: Finally, the enhanced feature $\overline{\mathbf{F}}$ is sliced back into the original slice format, resulting in $\{ \overline{\mathbf{P}}_i \}_{i=1}^N$, where $ \overline{\mathbf{P}}_i \in \mathbb{R}^{L \times D}$.

This process allows model to capture the complete semantics from high-resolution inputs while maintaining computational efficiency.

\subsection{Self-Mining Sampler}~\label{sec:sampler}
High-resolution images require processing more visual tokens, significantly increasing the computational load.
Existing solutions, such as Q-Former~\citep{li2023blip}, utilize a fixed number of learnable queries to compress visual features through a cross-attention mechanism. While effectively captures visual information regardless of image resolution in a computationally affordable manner, it suffers from several limitations:

\textbf{(i) Lacking positional information.} The learned queries lose positional information, degrading performance in tasks requiring spatial relationships and precise localization, such as visual reasoning.

\textbf{(ii) High training overhead.} Training Q-Former-like resamplers requires more data and longer training times to convert visual features into learnable queries, posing challenges in data-scarce domains.

To address these issues, we propose the self-mining sampler, as shown in \cref{fig:framework}~(b). The key idea of the self-mining sampler is to improve query initialization and reduce the receptive field that each query must compress based on the spatial priors. Specifically, we reshape the 1D output tokens of the vision encoder,~(\eg, CLIP-ViT), $\mathbf{P} \in \mathbb{R}^{L \times D}$, into 2D form, ${r \times r \times D}$, where $L = r \times r$. After applying average-pooling with kernel size $S \times S$, we obtain $\mathbf{P}^c \in \mathbb{R}^{r_2 \times r_2 \times D}$, where $r_2 < r$.
Next, we compute the final compressed tokens using the cross-attention mechanism by forcing the compressed token to perceive the $S \times S$ uncompressed tokens.
Unlike fixed learnable query-based methods, our self-mining sampler compresses the visual tokens based on themselves, preserving the original context and positional information while reducing training overhead.

\begin{figure*}[t!]
\begin{center}
\includegraphics[width=1\textwidth]{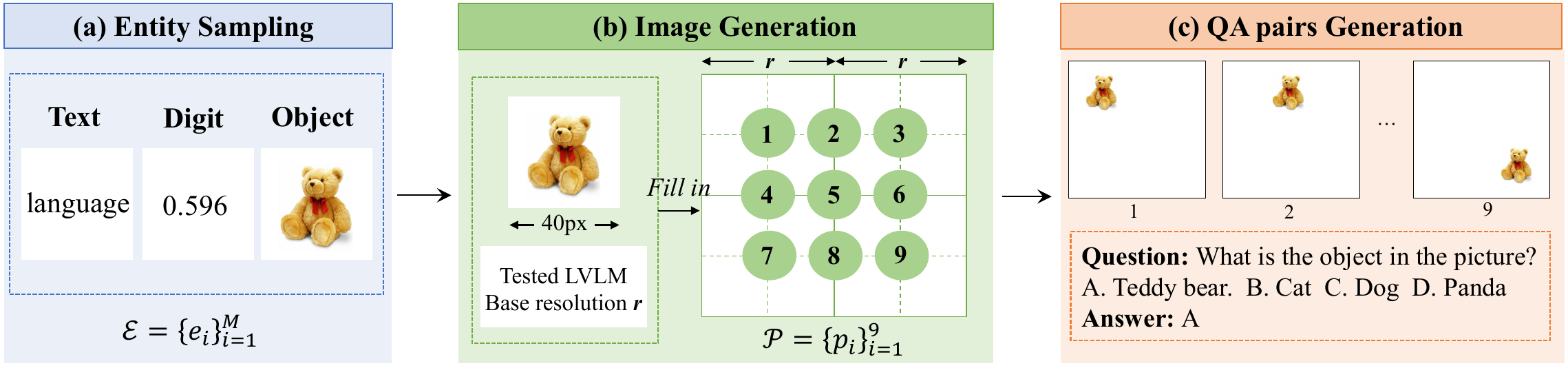}
\vspace{-2mm}
\caption{\textbf{Construction process of EntityGrid-QA benchmark.}
There are three steps: \textbf{(a) Entity Sampling.} Select one or two entities from the pre-defined entity set;
\textbf{(b) Image Generation.}  Put the selected entities in one position sampled from the nine pre-defined positions of the blank image, we can obtain the generated images. Note that the dash and solid lines in (b) are for illustration purposes only, and not presented to models.
\textbf{(c) QA pairs Generation.} Based on the generated images, entity category and positions, we can automatically generate the question-answer pairs (QAs).
}
\label{fig:benchmark}
\end{center}
\vspace{-4mm}
\end{figure*}

\subsection{EntityGrid-QA Benchmark}\label{sec:benchmark}
Existing benchmarks, particularly document-related datasets, can evaluate the fine-grained understanding of LVLMs. However, these benchmarks are inadequate for assessing the ability to handle fragmented inputs, as filtering slicing-related questions is time-consuming and labor-intensive. Therefore, we introduce a new benchmark named EntityGrid-QA, which is fully synthesized but still challenging for frontier models, to better assess LVLMs' ability to handle fragmentation.

\noindent\textbf{Construction process.} As shown in \cref{fig:benchmark}, the construction process of \dataset~consists of three main steps: Entity Sampling, Image Generation, and QA Pairs Generation.
Examples of our benchmark are provided in the Appendix.
Each step is detailed as follows:

\noindent~\textbf{(a) Entity sampling}. 
We first construct an entity set that includes various types such as English Words (\eg,  "apple"), Number (\eg, "0.596"), Object (\eg, a teddy bear) and Icon (\eg, "tomato") as shown in \cref{fig:benchmark}~(a).
Then, we select several entities from a predefined entity set, which can be denoted as 
$\mathcal{E}=\{ e_i \}_{i=1}^M$, where $e_i$ is the $i$-th entity and $M$ is the number of selected entities.

\noindent~\textbf{(b) Image generation}. The selected entities $\mathcal{E}$ are positioned in nine predefined positions (labeled 1 to 9) within a blank image using a 3x3 grid layout, as shown in \cref{fig:benchmark}~(b).
The resolution of the blank image is set to $2r \times 2r$, where $r$ is the base resolution of the pretrained vision encoder,~\eg, $224 \times 224$.
In this way, each image would be divided into four slices, and the each slice would match the input size of well-pretrained vision encoder, without the requirement of additional operations,~\eg, resize and padding.
Note that our \method~can process any number of slices, but some existing LVLMs,~\ie, LLaVA-Next~\citep{liu2023improved} can only receive four slices as input.
Hence, for a fair comparison, we only generate the images with a fixed resolution $2r \times 2r$.
For each entity $e_i$, we generate 9 images that iterate over all predefined 9 positions, with each position containing only one entity, as shown in \cref{fig:benchmark}.

\noindent~\textbf{(c) QA pairs generation}.
We mainly focus on evaluating the model's fine-granite recognition ability on the area of the slice boundary and center of the slices.
For each type of entity, we apply a specific question prompt,~\eg, \texttt{"What is the object in the picture?"}.
As shown in \cref{fig:benchmark}~(c), we formulate the question-answer pairs as the multi-choice problem.
Based on the selected entity $\mathcal{E}$ and the question $Q$, we apply the entity-specific augmentation to automatically generate the other three choices for the question. For example, given a number, the optional augmentations can be add, delete or shift the decimal point, or alter one of the digit of the number.
Note that for the triplets of image-question-answer of the same entity, it only varies in the position of the generated images while maintaining the same question, order of choices and ground truth answer which is perfectly assess the model.

After the construction, we create a training set of Entity-QA with 2k images covering 4 entity types and a testing set with 720 images and 20 entities per type. The entities in the training set and testing set are non-overlapped. Examples of the benchmark can be found in the Appendix. 

\noindent\textbf{Evaluation metric.} 
To evaluate the ability to handle fragmentation, we introduce a new metric that measures the precision discrepancies between entities located at the edge positions ($\mathcal{P}_\text{edge}=\{2,4,5,6,8\}$) and other locations ($\mathcal{P}_\text{center}=\{1,3,7,9\}$). We define:
\vspace{-1mm}
\begin{equation}
Acc_x = \frac{\sum_{p \in \mathcal{P}_x} A_p}{|\mathcal{P}_x|}, \quad x \in \{\text{edge, center}\}
\vspace{-1mm}
\end{equation}
where $A_p$ is the average accuracy when entities are located at position $p$, and $|\cdot|$ is the set size.

The Precision Discrepancies (PD) are defined as:
\vspace{-1mm}
\begin{align}
\text{PD}_1 = \frac{Acc_{\text{edge}}}{Acc_{\text{center}}},~~\text{PD}_2 = \frac{Acc_{\text{center}} - 
 Acc_{\text{edge}}}{Acc_{\text{center}}} \label{e:dis2}
\end{align}

\section{Experiment}
\subsection{Implementation Details}
\label{sec:implementation_details}
\definecolor{deemph}{gray}{0.6}
\newcommand{\gc}[1]{\textcolor{deemph}{#1}}

\definecolor{backcolor}{RGB}{232, 242, 255}

\begin{table*}[t]
\center
\tablestyle{2.5pt}{1}
\resizebox{\textwidth}{!}{
\begin{tabular}{llc|cccc|cc|ccc}
\toprule
&    &  & \multicolumn{4}{c|}{\it Document}   & \multicolumn{2}{c|}{\it Science} & \multicolumn{3}{c}{\it Comprehensive} \\
\multirow{-2}[0]{*}{Method}    &  \multirow{-2}[0]{*}{LLM}  &  \multicolumn{1}{c|}{\multirow{-2}[0]{*}{MaxRes}}  & \multicolumn{1}{l}{VQA-text} & \multicolumn{1}{l}{ChartQA} & \multicolumn{1}{l}{DocVQA} & \multicolumn{1}{l|}{InfoVQA} & \multicolumn{1}{l}{SQAI} & \multicolumn{1}{l|}{AI2D} & \multicolumn{1}{l}{MME} & \multicolumn{1}{l}{MMB} & \multicolumn{1}{l}{MM-Vet} \\
\midrule
\multicolumn{12}{c}{\textbf{\it General LVLMs (normal resolution)}} \\
\midrule
Qwen-VL-Chat & Qwen-7B & 448×448 & 61.5  & 66.3  & 62.6  & - & 68.2  & 57.7  & - & 60.6  & - \\
LLaVA-1.5 & Vicuna-1.5-13B & 336x336 & 61.3  & 18.2  & - & - & 71.6  & 59.5 & 1826  & 67.8  & 36.3 \\
LLaVA-MORE & Llama3.1-Ins-8B & 384x384 & 62.1  & - & - & - & 77.5  & 63.6  & 1846 & 73.1  & - \\
mPLUG-Owl3 & Qwen1.5-7B & 384x384 & 69.0 & - & - & - &  - & 73.4  & - & \textbf{77.6}  & 40.1 \\
\midrule
\multicolumn{12}{c}{\textbf{\it Document LVLMs}} \\
\midrule
DocPedia & Vicuna & 2560×2560 & 60.2  & 46.9  & 47.1  & 15.2  & - & - & - & - & - \\
UReader & Vicuna & ~896×1120 & 57.6  & 59.3  & 65.4  & 42.2  & - & - & - & - & - \\
TextMonkey+ & Qwen-7B & 896x896 & 64.3  & 59.9  & 66.7  & 28.6  & - & - & - & - & - \\
mPLUG-DocOwl2 & Qwen2-7B    & ~1512x2016 & 66.7  & 70.0  & 80.7  & \underline{46.4}  & - & - & - & - & - \\
\midrule
\multicolumn{12}{c}{\textbf{\it General LVLMs (higher resolution)}} \\
\midrule
Monkey & Qwen-7B & 896x896 & 67.6  & - & 66.5  & 36.1  & - & - & - & - & - \\
LLaVA-NeXT-8B & LLama3-Ins-8b & ~672x672 & 64.6  & 69.5  & 72.6  & - & - & 71.6  & \multicolumn{1}{l}{1603/-} & 72.1  & 41.7 \\
LLaVA-NeXT-13B & Vicuna-13B & ~672x672 & 67.1  & 62.2  & 70.9  & - & 73.6  & 70.0    & 1901  & 70.0    & \underline{48.4} \\
LLaVA-UHD & Vicuna-13B & ~672×1008 & 67.7  & - & - & - & 72.0 &  - & \multicolumn{1}{l}{1535/-} & 68.0    & - \\
Mini-Gemini-HD & Llama3-Ins-8b & 672x672 & 70.2  & 59.1  & 74.6  & - & 75.1  & 73.5  & \multicolumn{1}{l}{1606/-} & 72.7  & - \\
Cambrian-1-8B & Llama3-Ins-8B & 1024x1024 & 71.7  & 73.3  & 77.8  & - & \underline{80.4}  & 73.0    & \multicolumn{1}{l}{1547/-} & \underline{75.9}  & - \\
Cambrian-1-13B & Vicuna-1.5-13B & 1024x1024 & \underline{72.8}  & \underline{73.8}  & 76.8  & - & 79.3  & \underline{73.6}  & \multicolumn{1}{l}{1610/-} & 75.7  & - \\
\midrule
\rowcolor{backcolor}  \textbf{HiRes-LLaVA} & Llama3.1-Ins-8B & 1344x1344 & \textbf{74.2}  & \textbf{77.4}  & \textbf{84.9} & \textbf{55.7}  & \textbf{90.3}  & \textbf{74.9}  & \textbf{2213}  & 75.7 & \textbf{53.5} \\
\bottomrule
\end{tabular}
}
\vspace{-1mm}
\caption{\textbf{Quantitative results on 9 popular benchmarks.} `MaxRes' means the maximum resolution supported. `Document', `Science' and `Comprehensive' indicate the document-related VQA, Science VQA and comprehensive benchmarks.}
\vspace{-2mm}
\label{tab:results}
\end{table*}

\begin{table}[t]
 \centering
    \tablestyle{2.2pt}{1}
    \begin{tabular}{l|cc|cc|cc}
    \Xhline{1.2pt}
     { \textbf{Model}}
    & {\it \textbf{Acc$_{mean}$ $\uparrow$}} &  {\it \textbf{Acc$_{std}$  $\downarrow$}} &  {\it \textbf{Acc$_{e}$  $\uparrow$}} & {\it \textbf{Acc$_{c}$ $\uparrow$}} & {\it \textbf{PD$_1$ $\uparrow$}} & {\it \textbf{PD$_2$ $\downarrow$}} \\
    
  \midrule
   LLaVA-1.5 &  53.33 &   0.19 & 52.00 & 55.00 & 94.50 &    5.45  \\
   LLaVA-NeXT &  65.22 &   0.30 &  61.80 & 69.50 & 88.92 &    11.07         \\
 \midrule
 IXC-4KHD &  63.78 &  0.53 & 58.00 &  71.00  & 81.69 & 18.31  \\
\method & 70.20 &  0.19 & 68.40 & 72.50 & 94.34 & 5.60  \\
    \bottomrule
   \end{tabular}
    \vspace{-1mm}
      \caption{\textbf{Comparison with the state-of-the-art methods on \dataset.} `$\downarrow$' indicates lower scores are better, while `$\uparrow$' means the opposite. `Acc$_{mean}$' and `Acc$_{std}$', representing the mean and standard deviation of the average accuracy across three tasks. `Acc$_{e}$' and `Acc$_{c}$' show the average accuracy for entities at $\mathcal{P}_\text{edge}$ and $\mathcal{P}_\text{center}$, respectively. $PD_1$ and $PD_2$ are calculated using~\cref{e:dis2}. Note that IXC-4KHD and \method~are evaluated on 896x896 images and LLaVA-NeXT is evaluated on 672x672 images. The input resolution for LLaVA is 336px.}
    \label{tab:benchmark}
    \vspace{-5mm}
\end{table}

We utilize the CLIP-ViT-L/14-224px~\citep{radford2021learning} and InternViT-300M-448px~\cite{chen2023internvl} as the vision encoders, and Vicuna-v1.5-7B~\citep{chiang2023vicuna} and Llama-3.1-Instruction-8B~\citep{dubey2024llama} as LLM. We adopt a three-stage training approach, including an alignment stage, a capability enhancement stage and the instruction tuning stage. During the alignment stage, only the self-mining sampler is trainable. The learning rate is 1e-3. In the capability enhancement stage, we finetune the full model. The learning rate is 2e-5 for LLM and sampler, and 2e-6 for ViT. 
In the instruction tuning stage, ViT is frozen and the \adapter~is loaded with the LR of 2e-4. The learning rate of self-mining sampler and LLM is 2e-5. Four SliceRestore adapters are applied in the last four blocks of the vision encoder. All stages use the batch size of 256. 

We adopt AdamW~\citep{loshchilov2017adamw} as the optimizer with $\beta_1 = 0.9$ and $\beta_2=0.95$ to stabilize the training in the capability enhancement stage and the instruction tuning stage. In all stages, the learning rates are warmed up for the first 0.03 epochs and then adjusted by a cosine scheduler in the remaining training. We don't apply any weight decay in the training. The maximum number of slices is 9 for InternViT and 16 for CLIP-ViT.
Regarding the training data, we use the LLaVA-558k In the alignment stage, 1.8M long caption and OCR data in the capability enhancement stage and 3M multi-tasks data in the instruction tuning stage. 

\subsection{Experimental Setting}~\label{sec:setting}
We introduce experimental settings including the benchmarks and the compared LVLMs.

\noindent\textbf{Benchmarks.} We evaluate our models on (i) document-oriented VQA benchmarks, including VQA-Text~\citep{singh2019textvqa}, ChartQA test set~\citep{masry2022chartqa}, DocVQA test set~\citep{mathew2021docvqa}, InfoVQA test set~\citep{mathew2022infographicvqa}; (ii) general VQA benchmarks, including AI2D~\citep{kembhavi2016ai2d}, ScienceQA~\citep{lu2022scienceqa}; (iii) comprehensive benchmarks, including MMBench~\citep{liu2023mmbench}, MME~\citep{fu2023mme} and MM-Vet~\citep{yu2023mmvet}.

\noindent\textbf{Compared LVLMs.} We compare our model with SOTA LVLMs. (1) General LVLMs,~\ie , Qwen-VL~\citep{Qwen-VL}, LLaVA-1.5~\citep{liu2023improved}, LLaVA-MORE~\citep{cocchi2024llavamore}, mPLUG-Owl3~\citep{ye2024mplug}, Monkey~\citep{li2023monkey}, Mini-Gemini~\citep{li2024miniGemini}, LLaVA-UHD~\citep{xu2024llava}, LLaVA-NeXT~\citep{liu2024llavanext} and Cambrian-1~\citep{tong2024cambrian}. 
(2) Document LVLMs,~\ie, DocPedia~\citep{feng2023docpedia}, UReader~\citep{ye2023ureader}, TextMonkey~\citep{liu2024textmonkey} and mPLUG-Docowl2~\citep{hu2024mplug}.

\subsection{State-of-the-art Comparison}~\label{sec:sota}
\noindent\textbf{General benchmarks.} \Cref{tab:results} reports the performance comparison of our methods against state-of-the-art approaches on 11 benchmarks. Specifically, HiRes-LLAVA surpasses those general LVLMs with normal resolution inputs.
As for the document LVLMs with higher resolution inputs, HiRes-LLaVA demonstrates better performance on those document-related VQA benchmarks, for example, achieving 74.2 vs 66.7 of mPLUG-DocOwl2 on VQA-text, proving its capability to manage document-related tasks effectively.
Compared to Cambrian-1-13B that employs 4 vision encoders and is trained on 7M SFT data, our HiRes-LLaVA, with 8B LLM, one vision encoder and trained on 50\% less data than Cambrian-1, achieves better performance. These results indicate that HiRes-LLaVA has stronger generalization ability and robustness when dealing with complex documents, scientific problems, and comprehensive challenges.

\begin{figure*}[t]
    \begin{center}
    \includegraphics[width=1\textwidth]{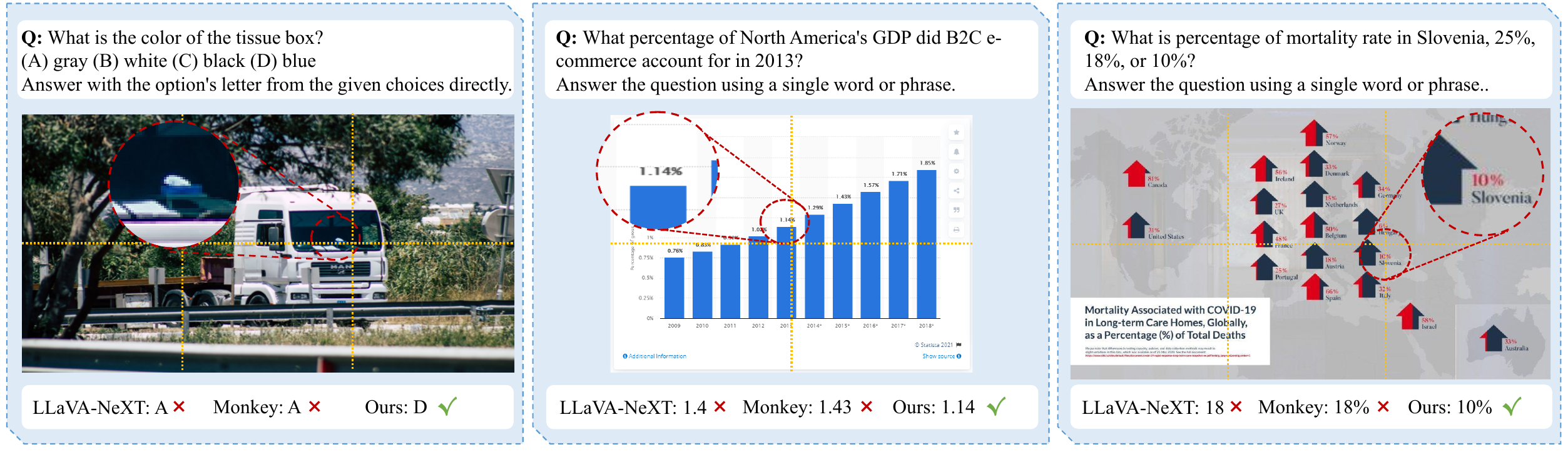}
    \vspace{-4mm}
    \caption{\textbf{Visualization comparison with the state-of-the-art methods.}
    Dash lines are only illustrated for the slice clarify.
    }
    \label{fig:visualization}
    \end{center}
    \vspace{-6mm}
\end{figure*}

\Cref{fig:visualization} shows a visual comparison of results generated by LLaVA-NeXT~\citep{liu2023improved}, Monkey~\citep{li2023monkey}, and our method, highlighting our superior performance, especially when the region of interest spans across slices.
For example, the number $1.14$ in \cref{fig:visualization}~(b) is split into two slices, causing Monkey to misidentify it as $1.4$. 
\begin{table*}[th]
 \centering
    \tablestyle{8pt}{1.05}
    \begin{tabular}{lcc|ccccc|ccc}
    \Xhline{1.2pt}
    \rowcolor{mygray}
    \multicolumn{3}{c|}{\it \textbf{Components}}
    &\multicolumn{5}{c|}{\it \textbf{Document}}
    &\multicolumn{3}{c}{\it \textbf{Comprehensive}}
    \\
    \rowcolor{mygray}
      Downsampler  &  SRA  & Separator & VQA-Text &  ChartQA &  DocQA &  InfoVQA &  Avg. &   MMB  &  MM-Vet & MME-P    \\
      \midrule
      \multicolumn{3}{l|}{\textbf{Baseline(LLaVA)}}  & 53.3 & 23.8 & 22.6 & 26.0 & 31.4 & 64.0 & - & 1424.7 \\
      \midrule
      ConcatChannel  & \xmarkg & \xmarkg & 60.3 &  54.4 &	54.8 &	34.3 &   50.9 &  60.8    &  30.2 & 1355.5   \\
      Resampler &  \xmarkg  & \xmarkg & 58.8 & 49.8 &  42.8  &  32.6 & 46.0 & 59.6 & 26.6 & 1404.0   \\ 
      C-Abstractor & \xmarkg  &  \xmarkg & 59.0 & 55.6 & 54.7  &  36.7 &   51.5  &  63.5  & 30.4 	& 1393.5  \\
      SMS & \xmarkg  & \xmarkg & 60.0 & 56.2 & 58.0 &  37.4 & 52.9 &  63.3 & 31.1 & 1411.3  \\
      SMS & G & \xmarkg & 60.9 & 56.2 & 57.2 & 38.2 & 53.1 & {65.5} &  30.6  & {1415.8}   \\
      SMS & G \& L & \xmarkg & {61.5} & {56.9} & {57.6} & {38.4} & {53.6}  &  {64.9}  & {33.8} & {1452.9} \\  
      SMS & G \& L &  \checkmark & 61.8 & 58.8 & 59.7 & 41.4 & 55.4 & 65.5 & 33.8 & 1456.1 \\
      \midrule
       \multicolumn{3}{l|}{improvement relative to the baseline}  & \improve{+8.5} & \improve{+35.0} & \improve{+37.1} & \improve{+15.4} & \improve{+24.0 }& \improve{+1.5} & - & \improve{+31.4} \\
    \bottomrule
    \end{tabular}
\vspace{-1mm}
 \caption{\textbf{Ablation study of different proposed modules.} Note that `G', and `G-L' indicate using the global fusion and the combination of them respectively. All results are conducted with the maximum number of slices is 16 except the baseline model, LLaVA. The last row is the improvement over the baseline model.}
\label{tab:effectofmodule}
\vspace{-3mm}
\end{table*}

Our method, with the SRA capturing complete global high-resolution information, correctly predicts the answers.

\noindent\textbf{\dataset.} 
To evaluate the ability to address input fragmentation, we compare LLaVA-1.5 with normal resolution input and two SOTA slicing-based LVLMs. The results are presented in \cref{tab:benchmark}.
According to the experimental results, we can observe two key findings:
(i) Slicing the high-resolution image will bring the fragmentation issue. Although LLaVA-NeXT performs better than LLaVA-1.5 on both $Acc_e$ and $Acc_c$, it suffers significantly from fragmentation, as indicated by a 5.58\% drop in $PD_1$ and a 5.62\% increase in $PD_2$.
(ii) Our method, utilizing SRA, significantly outperforms SOTA LVLMs in handling entities at the edges of slices. For example, IXC-4KHD (InternLM-Xcomposer-4KHD)\citep{zhang2023internlm} exhibits a notable discrepancy between $Acc_{e}$ and $Acc_{c}$, with scores of $58.0\%$ and $71.0\%$, respectively. In contrast, our method achieves higher accuracy at both the edges and the center of the slices ($68.4\%$ for $Acc_{e}$ and $72.5\%$ for $Acc_{c}$) and also obtains a smaller difference, with $94.34\%$ for $PD_1$ and $5.6\%$ for $PD_2$, which is close to the LVLMs with normal resolution inputs.


\begin{figure*}[t]
\begin{center}
\includegraphics[width=0.9\textwidth]{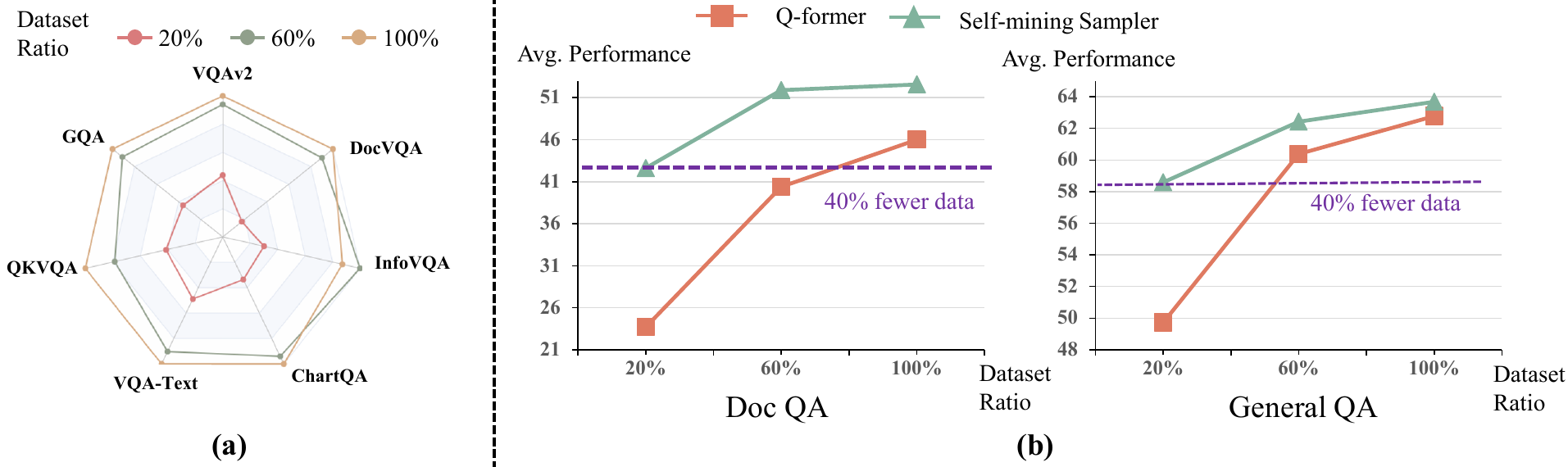}
\vspace{-3mm}
\caption{\textbf{(a) Ablation on data efficiency of \method.} 
We sample the training data mixture at ratios of 20\%, 60\%, and 100\% and report the performance of our \method~on seven benchmarks.
\textbf{(b) Data efficiency comparison with Q-former and our proposed self-mining sampler (SMS).} The performance on `Doc QA' is averaged from DocVQA, ChartQA and InfoVQA. The performance on `General QA' is averaged from the other four benchmarks. Our SMS can use $40\%$ fewer data to achieve competitive performance compared with Q-former, indicating our method's efficiency. Note that both Q-former and our SMS apply one cross-attention block.
}
\label{fig:data-efficient}
\end{center}
\vspace{-5mm}
\end{figure*}

\subsection{Ablation Study}~\label{sec:ablation}
In this section, we conduct ablation studies to evaluate the effect of our proposed modules. In our ablation study, we conduct the experiments following LLaVA's setting on the LLaVA 1.2M data~\citep{liu2023improved} with additional 79K document-oriented data, which is essential to evaluate the high-resolution LVLMs, in the instruction tuning stage, i.e., DocVQA~\citep{mathew2021docvqa}, ChartQA~\citep{masry2022chartqa} and InfoVQA~\citep{mathew2022infographicvqa}. Unless specified, we use LoRA~\citep{hu2021lora} to efficiently finetune pretrained LLM, \ie, Vicuna-1.5-7B and CLIP-ViT-Large-224px as the vision encoder with maximum 16 slices in our ablation. 
\begin{table}[t!]
 \centering
    \begin{adjustbox}{max width=1.0\linewidth}
    \tablestyle{2pt}{1.05}
    \Large
    \begin{tabular}{cc|cccccc}
    \Xhline{1.2pt}
\toprule
 $M$ & \#Tokens &  VQA-Text &  ChartQA &  DocQA &  InfoVQA  &  MMB  & MME-P  \\
  \midrule
  4 & $\sim$320 &  56.2 & 42.5 & 37.0 & 28.8 &  65.1 & 1436.3  \\ 
  9 & $\sim$640 &  59.9 & 51.6 & 49.3 & 34.9 & 64.3 & 1450.0 \\ 
  16 & $\sim$1088 & 61.8 & 58.8 & 59.7 &  41.4 & 65.5 & 1456.1  \\
  \bottomrule
  \end{tabular}
  \end{adjustbox}
  \vspace{-2mm}
 \caption{\textbf{Effect of different numbers of slices.} $M$ and `\#Tokens' indicate the maximum number of slices and visual tokens in the high-resolution images, respectively.}
  \label{tab:slices}
  \vspace{-4mm}
\end{table}

\noindent \textbf{Effect of the proposed modules.} 
We ablate the two main components of our \method, specifically the \adapter~(SRA) and the self-mining sampler (SMS), as shown in \cref{tab:effectofmodule}. Our findings are as follows:
Our SMS demonstrates superior performance compared to other samplers, notably outperforming Resampler~\citep{Qwen-VL} by $6.9\%$ on the average score across four benchmarks.
Integrating the model with SRA leads to further improvements across these benchmarks.
Additionally, the introduction of learnable queries to isolate slice representations, referred to as Separator, results in a $1.8\%$ enhancement in the average score.

\noindent \textbf{Ablation study of kernel sizes in SMS.} Here we conduct the ablation study of the self-mining sampler.
In \cref{tab:sampler}, we compare the performance of the average pooling with different kernel sizes,~\ie, $S \times S$ in \cref{sec:sampler}.
The results show that as the kernel size increases,~\ie, the fewer visual tokens, the performance would degrade, since the information loss.

\noindent \textbf{Ablation study of the number of high-resolution image slices.} As shown in \cref{tab:slices}, the number of slices significantly affects the model's performance on the document-related benchmarks. Specifically, when increasing the number of slices from 4 to 16, the average performance improves by 14.3\% on four document-related benchmarks. As for the comprehensive benchmarks, larger number of slices doesn't effect model's performance on MMBench too much and can bring a 19.8 improvement on MME-Perception.
\begin{table}[t]
 \centering
    \begin{adjustbox}{max width=\linewidth}
    \tablestyle{2pt}{1.05}
    \begin{tabular}{ccrccccc}
    \Xhline{1.2pt}
   \textbf{Base} & \textbf{Kernel} & {\small \textbf{Max \#Tokens}} & \multicolumn{5}{c}{\small  \it \textbf{}}  \\
     \textbf{ Res.} & \textbf{Size}  & (Token/Slice)  &   VQA-Text &  ChartQA &  DocVQA &  InfoVQA &  Avg. \\
  \midrule
   224 & $2 \times 2$ & 1088 (64) & 61.8 & 58.8 & 59.7 & 41.4 & 55.4  \\
   224 & $4 \times 4$ & 272 (16) &  59.6 & 53.9 &  46.3  & 33.0 & 48.2  \\
   224 & $8 \times 8$ & 68 (4) & 54.9 & 46.8 & 35.3 & 29.6  & 41.7   \\
    \midrule
   336 & $2 \times 2$ & 2448~(144) & 63.6&58.5&65.7&40.7&57.1 \\
   336 & $3 \times 3$ & 1088~(64) & 61.2&56.7&59.8&38.7&54.1 \\
   336 & $4 \times 4$ & 512~(36) &  61.4&53.3&54.3&34.3&50.8 \\
    \bottomrule
   \end{tabular}
    \end{adjustbox}
    \vspace{-2mm}
     \caption{\textbf{Effect of different downsample kernel sizes in the self-mining sampler.} `Kernel Size' is $S \times S$ defined in~\cref{sec:sampler}. `Base Res.' indicates the base resolution of the vision encoder. `Max \#Tokens' indicates the maximum number of visual tokens,~\ie, $M \times r_2 \times r_2$, as the maximum number of slices $M$ is 16.}
    \vspace{-2mm}
    \label{tab:sampler}
\end{table}

\noindent \textbf{Ablation study of the selection of vision encoder and language model.} In \cref{tab:ablation_llm_vit}, we evaluate the performance of different vision encoders and large language models on LVLM Benchmarks. Experimental results show that compared to Vicuna-1.5-7B, LLaMA3.1-8B-Instruct can significantly improve the model's performance on both document-related benchmarks and comprehensive benchmarks. Additionally, InternViT-300M-448px can maintain performance on comprehensive benchmarks and further improve all document-related benchmarks by increasing the base resolution and the number of visual tokens.

\begin{table}[t!]
 \centering
    \tablestyle{1.5pt}{1.05}
    \begin{adjustbox}{max width=1\linewidth}
    \begin{tabular}{llcccccc}
    \Xhline{1.2pt}
  Vision Encoder & LLM & VQA-Text &  ChartQA &  DocQA &  InfoVQA &   MMB  & MME-P    \\
  \midrule
  CLIP-ViT-224px & Vicuna & 61.8 & 58.8 & 59.7 & 41.4 &  65.5 & 1456.1  \\
  CLIP-ViT-224px & Llama3.1 &  60.5 & 58.6 & 67.2 & 47.2 & 68.1 & 1453.4    \\
  InternViT-448px & Llama3.1 & 63.4 & 65.9	& 74.4 & 53.2 & 68.0 &  1459.1    \\
\bottomrule
\end{tabular}
\end{adjustbox}
\vspace{-2mm}
 \caption{\textbf{The ablation study of different vision encoder and large language models. } Note that CLIP-ViT-224px uses 16 maximum slices and InternViT-448px uses 9 slices.}
\label{tab:ablation_llm_vit}
\vspace{-2mm}
\end{table}

\noindent \textbf{Data efficiency analysis.}
We evaluated the data efficiency of our method, \method, by subsampling the training data mixture at ratios of 20\%, 60\%, and 100\%. Results in \cref{fig:data-efficient}~(a) show that using the entire dataset achieves optimal performance. Remarkably, with only 60\% of the data, performance remains above 90\% of the full dataset's level, highlighting the potential for improved data efficiency.
Additionally, we compared our self-mining sampler's efficiency against the commonly used Q-former in LVLMs. As depicted in \cref{fig:data-efficient}~(b), our method performs competitively with Q-former even with only 20\% of the data, demonstrating its effectiveness and efficiency.

\section{Conclusion}\label{sec:conclusion}
In this paper, we present \method, a large visual-language model (LVLM) designed to efficiently address input fragmentation caused by current slicing-based high-resolution LVLMs. To evaluate this capability, we introduce a new benchmark, \dataset, which focused on identification tasks on various entities. Comprehensive experimental results on 9 popular existing benchmarks and \dataset~demonstrate the effectiveness of \method. Analytical evaluation and visualization results are provided for a deeper understanding of the model's performance.

\section*{Acknowledgements} 
We gratefully acknowledge supports of MindSpore, CANN (Compute Architecture for Neural Networks) and Ascend AI Processor used for this research.
This work is supported by National Key Research and Development Program of China (2024YFE0203100) , Shenzhen Science and Technology Program No.GJHZ20220913142600001, National Natural Science Foundation of China (NSFC) (No.62476293, 62441615 and 62201484), Nansha Key R\&D Program under Grant No.2022ZD014 and General Embodied AI Center of Sun Yat-sen University.

{
\small
\bibliographystyle{ieeenat_fullname}
\bibliography{main}
}

\clearpage
\appendix
\section*{Appendix} 
\section{Implementation Details}
\label{sec:appendix_implementation_details}

\noindent \textbf{Training datasets.}
\Cref{tab:pretrain_data} shows the detailed dataset construction of the capability enhancement stage of \method. Specifically, it has 830K captioning including the ShareGPT4V~\citep{chen2023sharegpt4v}, ShareGPT4o~\citep{sharegpt4o} and
 ALLAVA~\citep{chen2024allava}. There are 821K OCR data from SynthDoG~\citep{kim2022synthdog} including English OCR data as well as MMC-Alignment~\citep{liu2023mmcinst}, UReader~\citep{ye2023ureader}, K12 printed~\citep{k12} which is a short OCR dataset. There is also 200K text instruction data from Magpie Pro~\citep{xu2024magpie}, sampling from the data generated by Llama3.1-70B, Llama3-70B, and Qwen2-72B.

\begin{table}[h]
    \centering
    \renewcommand\arraystretch{1.0} 
    \setlength{\tabcolsep}{1.1mm}    
    \begin{tabular}{p{1.5cm} p{4cm} p{2cm}}
         \toprule
        \textbf{Task} &  \textbf{Datasets(\# Sample)} & \textbf{Sum} \\
         \midrule
         \textbf{Caption} &  ShareGPT4V(89k),\newline 
                    ALLAVA4V(684k),\newline
                    ShareGPT-4O(57k).  &  830K(44.8\%)  \\
         \midrule
          \textbf{OCR}    & SynthDoG-EN(300k), \newline
                MMC-Alignment(200k),\newline
                UReader(101k),\newline
                K12 printed(120k),\newline
                SynthDoG-ZH(100k).  & 821k(44.4\%) \\
        \midrule
        \textbf{Text} & Magpie Pro(200k) &  200k(10.8\%) \\
         \midrule
      \textbf{Total} &    &  \textbf{1.8M}    \\
         \bottomrule
    \end{tabular}
    \caption{Datasets in the capability enhancement stage.} 
    \label{tab:pretrain_data}
\end{table}

\begin{table*}[th]
\renewcommand{\arraystretch}{1.2}
\setlength\tabcolsep{3pt}
\centering
\begin{tabular}{p{3cm} p{8cm} p{2cm}}
\toprule
\textbf{Task} &  \textbf{Datasets(\# Sample)} & \textbf{Sum} \\
\midrule
\textbf{{General QA}} & LLaVA(135K), ALLaVA(660K) VQAv2(83K), \newline 
                    GQA(72K), OKVQA(9K), A-OKVQA(66K), \newline VSR(12K), ShareGPT4V(89K), 
                    TextCaps(22K), Laion-GPT4V(11K), ShareGPT-4O(57K), 
                    RAVEN(3K), Visual7w(14K), RefCOCO(48K), VG(86K) & 1.4M (48.0\%) \\
                    
\midrule
\textbf{{Science}}  & ScienceQA(19K), ai2d(14K), ViQuAE(4K), \newline
                    TextbookQA(21K), IconQA(30K), \newline
                    Data Engine(50K)  &   139K(4.6\%)               
\\
\midrule
\textbf{Doc QA/OCR  }    &    OCRVQA(80K), TextVQA(57K), SynthDog(30K), \newline
             LLaVAR(39K), WikiTableQuestions(29K), \newline KleisterCharity(15K), iiit(6K),  MLHME(30K),  \newline
             VisualMRC(19K),
             ChartQA(48K), DocVQA(102K), \newline InfoVQA(33K), DVQA(200K),
             PlotQA(10K), \newline
             TAT-DQA(2K), TableFact(65K), WebSRC(5K)\newline
             DeepForm(8K), Chart2text(27K)  \newline
        Vistext(10K),
        chrome writting(9K),
        IAM(6K),  \newline
        Rendered text (10K), 
        Orand-CAR-A(2K), 
        lrv-chart(2K),  \newline
        ROBUT-SQA(9K),
        ROBUT-WTQ(4K),
        Hitab(3K),  \newline
        Diagram-image-to-text(0.3K). & 0.9M(30.1\%)  \\
\midrule                      
\textbf{Code Generation}   &  WebSight(50K)  & 50K(1.7\%)  \\ 
\midrule
\textbf{Text-only }  & Magpie-Pro(150K), 
                    Evol(142K), \newline
                    mathinstruct(81K),
                    mathplus(95K).  &  469K(15.6\%)         \\
\midrule
\textbf{Total}      &    &  \textbf{3M }       \\
\bottomrule
\end{tabular}
\caption{Summary of datasets used in the instruction tuning stage.}
\label{tab:sft_dataset}
\end{table*}

\Cref{tab:sft_dataset} shows the detailed construction of the 3M instruction tuning dataset. First, we remove 23K caption data and ShareGPT data from original LLaVA-158K~\citep{liu2023llava} and include GPT4V/GPT4o-generated caption data, i.e., LAION-GPT4v~\citep{laion_gpt4v_dataset}, ShareGPT4V~\citep{chen2023sharegpt4v}, ShareGPT4o~\citep{sharegpt4o} and ALLAVA instruction data~\citep{chen2024allava}. To enhance the common knowledge of our model, we convert the visual spatial reasoning~\citep{liu2023vsr}, AI2D~\citep{kembhavi2016ai2d}, and Science QA~\citep{lu2022scienceqa} training set into the instruct-tuning data.
To activate the understanding science, we collect data from ViQuAE~\citep{lerner2022viquae},
TextbookQA~\citep{kembhavi2017tqa}, IconQA~\citep{lu2021iconqa} and
sampled 50k data from the Cambrian's Data Engine~\citep{tong2024cambrian}.                 
We also collect document-oriented data from diverse datasets, includes ChartQA~\citep{masry2022chartqa}, DVQA ~\citep{kafle2018dvqa}, PlotQA ~\citep{methani2020plotqa}, OCRVQA~\citep{mishra2019ocrvqa}, ST-VQA~\citep{biten2019stvqa}, DocVQA~\citep{clark2017docqa}, InfoVQA~\citep{mathew2022infographicvqa}, DeepForm~\citep{svetlichnaya2020deepform}, TAT-DQA~\citep{zhu2022tatdqa}, TableFact~\citep{chen2019tabfact},  LRV-Chart\citep{liu2023aligning} and WebSRC~\citep{chen2021websrc}. We merge some datasets from Cauldron~\citep{laurençon2024cauldron}, including RAVEN, ROBUT-SQA, ROBUT-WTQ, HiTab, IAM, Rendered Text, ORAND-CAR-A, Visual7W, Chart2Text, AI2D, vistext, Diagram-image-to-text. 

\noindent \textbf{Module Design Details.} The self-mining sampler consists of one cross-attention block with an output layer norm. The cross-attention block has a cross-attention layer and a FFN. Both of them apply the residual shortcut. The cross-attention layer has two layer norm for the query and key/value, respectively. As for the SliceRestore Adapter, the parameters of the self-attention layer with the layer norm are initialized from the pretrained CLIP self-attention at the same depth. To provide the positional information between slices, we apply a 2D RoPE~\citep{su2024roformer,sun2023evaclip} on the global fusion module.

\noindent \textbf{Training pipeline.} We list the hyperparameters for the  three-stage training at \cref{tab:training_pipeline}. 
\begin{table*}[t]
  \centering
  \renewcommand{\arraystretch}{1.2}
  \resizebox{0.9\textwidth}{!}{%
  \begin{tabular}{ll|c|c|c}
  \toprule
  & \textbf{Settings} & \textbf{Stage-1} & \textbf{Stage-2} & \textbf{Stage-3} \\
  \midrule
  \multirow{2}{*}{\rotatebox[origin=c]{90}{\footnotesize \textit{Vision}}} & \textbf{Resolution} & 448\footnotesize{$\times$\{\{1$\times$2\}, $\cdots$, \{3$\times$3\}\}}   & 448\footnotesize{$\times$\{\{1$\times$2\}, $\cdots$, \{3$\times$3\}\}} & 448\footnotesize{$\times$\{\{1$\times$2\}, $\cdots$, \{3$\times$3\}\}} \\
  & \# Tokens & Max $256 \times (1 + 9)$ & Max $256 \times (1 + 9)$   & Max $256 \times (1 + 9)$  \\
  \midrule
  \multirow{2}{*}{\rotatebox[origin=c]{90}{\footnotesize \textit{Data}}} & \textbf{Dataset} & LLaVA-Pretrain & Enhancement (\cref{tab:pretrain_data}) & SFT (\cref{tab:sft_dataset}) \\
  & \# Samples & 558K & 1.8M & 3M \\
  \midrule
  \multirow{5}{*}{\rotatebox[origin=c]{90}{\footnotesize \textit{Training}}} & \textbf{Trainable} & Projector & ViT \& Projector \& LLM & SRA \& Projector \& LLM \\
  & \textbf{Load SRA} & \xmark & \xmark & \cmark \\
  & \textbf{Batch Size} & 256 & 256 & 256 \\
  & \textbf{LR: LLM} & $2\times 10^{-5}$ & $2\times 10^{-5}$ & $2\times 10^{-5}$ \\
  & \textbf{LR: Projector}  & $1\times 10^{-3}$ & $2\times 10^{-5}$ & $2\times 10^{-5}$ \\
  & \textbf{LR: ViT / SRA} & - & $2\times 10^{-6}$ & $2\times 10^{-4}$ \\
  & \textbf{Epoch} & 1 & 1 & 1 \\
  \bottomrule
  \end{tabular}
  }%
  \vspace{-1mm}
  \caption{
  \textbf{Detailed configuration for three-stage training of \method.}
  The table illustrates the vision configurations, dataset characteristics, and training hyperparameters.
  }
  \label{tab:training_pipeline}%
\end{table*}%

\noindent \textbf{Evaluation details.} We utilize the open-source evaluation tools, lmms-eval~\citep{lmms_eval2024}, to align our evaluation method to LLaVA-NeXT~\citep{liu2024llavanext}. 

\noindent \textbf{Benchmark construction.} In our EntityGrid-QA, the construction of multiple choices is a vital part of EntityGrid-QA. For different types of entities, we apply different augmentations to obtain the other three choices for each question. For text and decimal, we randomly delete, add, or change one letter or digit. The object figures are collected from the COCO dataset~\cite{lin2014microsoft}. For both categories of icons and objects, we use GPT-4 to list three other entities' names with similar appearance as the negative options.

\section{More Ablation}
\noindent \textbf{Comparison on the Same Training Set}
To demonstrate the effectiveness of our method, we compare the performance of LLaVA-1.5 and our method trained on the same data.
Specifically, we train these two models on two different scale training data set,~\ie, LLaVA-655K~\citep{li2023llava} and LLaVA-655K with additional Doc-79K data (the dataset of our ablation setting).
Results from \cref{tab:same_training_data} show that adding 79K document data can highly improve models' performance on ChartQA, DocQA and InfoVQA but slightly drops the performance on MMBench and MME-Perception. Hires-LLaVA outperforms the LLaVA-1.5 under these two training data sets, confirms that the superior performance can be attributed to the method itself rather than the volume of data. 

\noindent \textbf{Ablation of the separators}
To further evaluate the effect of the separators, we conduct experiments on whether the separators are different or the same. \cref{tab:seperator} demonstrates that using separated separators greatly outperforms using the same ones which would confuse the model about the position of slices.

\section{Efficiency Analysis}
\paragraph{Comparison with other LVLMs.} To validate the efficiency of our method, we compare the computational cost, training, and inference times with various LVLMs in \cref{tab:efficency}. For computational cost, we report the FLOPs of the ViT backbone, connector, and LLM components for each model. Experimental results demonstrate that \method, despite processing inputs at twice the resolution of LLavA-Next (1344$^2$ vs. 672$^2$), is able to reduce training time by approximately 74\%.
\begin{table}[t]
     \centering
    \tablestyle{1.2pt}{1.05}
        \begin{adjustbox}{max width=\linewidth}
        \begin{tabular}{ll|cccccccc}
        \Xhline{1.5pt}
      Model &  Data &  VQA-Text &  ChartQA &  DocQA &  InfoVQA &   MMB &  MME-P \\
      \midrule
      LLaVA-1.5 &  LLaVA-665k   & 53.3 & 13.7 & 14.2 & 19.4  & 71.1 & 1459.66  \\ 
      LLaVA-1.5 &  LLaVA-665k + Doc-79k & 53.3 & 23.8 & 22.6 & 31.4  & 70.7 & 1424.6 \\ 
      \midrule
      HiRes-LLaVA &  LLaVA-665k & 62.4 &	19.8 & 37.7 & 26.0 & 72.3 & 1486.1 \\
      HiRes-LLaVA &  LLaVA-665k + Doc-79k & 62.3 & 57.6 &  58.5 &  39.2 & 71.1 & 1444.8  \\
      \bottomrule
      \end{tabular}
      \end{adjustbox}
    \vspace{-3mm}
    \caption{Ablation study of different training data. Using the same training data, our \method~consistently outperforms LLaVA-1.5, demonstrating the superior effectiveness of our approach.}
      \label{tab:same_training_data}
\end{table}

\paragraph{Comparison with other downsampling methods.} We also compare the FLOPs and training time of our proposed downsampling strategy SMS with other vision token downsamplers, including ConcatChannel~\citep{chen2023minigptv2}, Q-Former~\citep{Qwen-VL}, and C-Abstractor~\citep{cha2023honeybee}, as shown in \cref{tab:individual_components}. The results show that our SMS, even when combined with additional components like SRA, achieves competitive efficiency compared to existing state-of-the-art downsamplers.

\begin{table}[t]
 \centering
 \small
    \begin{adjustbox}{max width=0.9\linewidth}
    \tablestyle{2pt}{0.65}
    \Large
    \begin{tabular}{c|cccccc}
\toprule
 Type &  VQA-Text &  ChartQA &  DocQA &  InfoVQA  &  MMB  & MME-P  \\
  \midrule
  Same & 57.2 & 39.7 & 52.6 & 37.6 & 61.3 & 1379.8   \\ 
  Separated & \textbf{61.8} & \textbf{58.8} & \textbf{59.7} & \textbf{41.4} & \textbf{65.5} & \textbf{1456.1}  \\ 
  \bottomrule
  \end{tabular}
  \end{adjustbox}
  \caption{Ablation of the separator. `Separated` means three separators are the difference and `Same` means that three separators are the same.}
  \label{tab:seperator}
\end{table}

\begin{table}[t!]
    \centering
    \tablestyle{2.5pt}{1.05}
    \begin{adjustbox}{max width=1.\linewidth}
    \begin{tabular}{c|c|ccc|cc}
    \toprule
    \textbf{Training} & \textbf{Inference} & \multicolumn{3}{c|}{\small \it \textbf{FLOPs}} & \textbf{Training}  &  \textbf{Inference}\\
    \textbf{batch size} & \textbf{Resolution} & ViT  & Connector &  LLM & \textbf{ time} &  \textbf{ time}   \\
    \midrule
    \multicolumn{7}{c}{  \it \textbf{HiRes-LLaVA}} \\
    \midrule
    2 &  1344x1344   & 6.6 T & 195.2 G & 37.1 T & 60.7h \footnotesize\textbf{\textcolor{purple}{(15.9\%)}}  & 15.4m \\
    \midrule
    \multicolumn{7}{c}{  \it \textbf{ HiRes-LLaVA w/o SRA }} \\
    \midrule
    2 &  1344x1344  & 6.5 T & 195.2 G & 37.1 T & 59.5h  \footnotesize\textbf{\textcolor{purple}{(15.6\%)}} & 12.9m \\
    \midrule
    \multicolumn{7}{c}{  \it \textbf{  LLaVA-Next (LLaVA-1.6) }} \\
    \midrule
    2 &1344x1344  &  \multicolumn{5}{c}{\textcolor{red}{Out of the memeory}}\\
    \midrule
    1 & 672x672  & 1.9 T	& 120.8	G & 44.0 T & 381.0h & 13.2m  \\
    \bottomrule
    \end{tabular}
    \end{adjustbox}
    \label{tab:efficency}
    \caption{Comparison of the efficiency of different models. Note that training time is assessed under the SFT setting on a machine with 8 V100 GPUs. The inference time is assessed on the InfoVQA benchmark with 6096 images by using the lmms-eval. Note that using the same batch size per device and resolution, LLaVA-Next would be out of the memory. The ratios of training time for ours relative to LLaVA-Next are marked in \textbf{\textcolor{purple}{purple}}.
    }
\end{table}

\begin{table}[h!]
 \centering
    \begin{adjustbox}{max width=1\linewidth}
    \begin{tabular}{lc|ccc|c}
    \Xhline{1.3pt}
    \multicolumn{2}{c|}{\small  \it \textbf{Components}}
    &\multicolumn{3}{c|}{\small  \it \textbf{FLOPs}}
    & Training  \\
  Downsampler  &  SRA  &  ViT  & Sampler  &  LLM  &  Time  \\
  \midrule
  NoDownsample  & \xmarkg  &  6.5 T &  410.8 G & 148.3T & - \\
  \midrule
  ConcatChannel  & \xmarkg  &  6.5 T & 164.3 G & 37.1 T & 58.6h  \\
  Q-Former &  \xmarkg &  6.5 T  & 205.5 G & 37.1 T & 58.9h  \\ 
  C-Abstractor & \xmarkg &  6.5 T & 258.2 G & 37.1 T & 60.7h  \\
  SMS & \xmarkg & 6.5 T & 195.2 G & 37.1 T & 59.5h  \\ 
  SMS & \checkmark &  6.6 T & 195.2 G & 37.1 T & 60.7h \\
    \bottomrule
   \end{tabular}
    \end{adjustbox}
      \caption{Ablation study of the efficiency of individual components for different downsamplers. We assume the inputs are an image with 16 slices and 100 text tokens. Note that no downsampling method causes out-of-memory (OOM) issues during training. Training time is assessed under the SFT setting on a machine with 8 V100 GPUs.}
     \label{tab:individual_components}
\end{table}
\begin{table}[h]
    \centering
    \begin{adjustbox}{max width=0.9\linewidth}
    \begin{tabular}{lll}
         \toprule
    Benchmarks & Slicing Strategy  &  Target Issue  \\
     \midrule
    LLaVA-UHD's & Overlapped  & Counting \\
    Our EntityGrid-QA & Non-overlapped  & Fragmentation \\
    \bottomrule
    \end{tabular}
    \end{adjustbox}
    \caption{The differences between our EntityGrid-QA and LLaVA-UHD's benchmark~\cite{xu2024llava}.} \label{tab:appendix_compared_llava_uhd}
\end{table}
\section{Discussion}
\noindent \textbf{What's the goal of the EntityGrid-QA benchmark?} The goal of our EntityGrid-QA benchmark is to assess the fragmentation issue in LVLMs (Large Vision-Language Models) when processing high-resolution inputs, rather than their ability to identify different types of objects. To address this, EntityGrid-QA synthesizes images by iteratively placing objects in different positions, allowing us to evaluate how these models perform on the edges and the center of the slices. Compared to harvesting real-world images with answer targets on the edges of slices, the synthesized approach is more simple-to-collect, effective, flexible, sufficient to evaluate the fragmentation issue.

\noindent \textbf{Compared with LLaVA-UHD.} The target issues and slicing strategies are different between Hires-LLaVA and LLaVA-UHD~\cite{xu2024llava}. While LLaVA-UHD reveals the counting problem in the overlap slicing strategy for the high-resolution image inputs, Hires-LLaVA focuses on the fragmentation issues of non-overlapped slicing strategy which is commonly used in recent open-sourced high-resolution LVLMs. \Cref{tab:appendix_compared_llava_uhd} summarize the differences of our EntityGrid-QA and LLaVA-UHD's benchmark.

\section{More Visualization}
\begin{figure*}[t!]
\begin{center}
\includegraphics[width=1\linewidth]{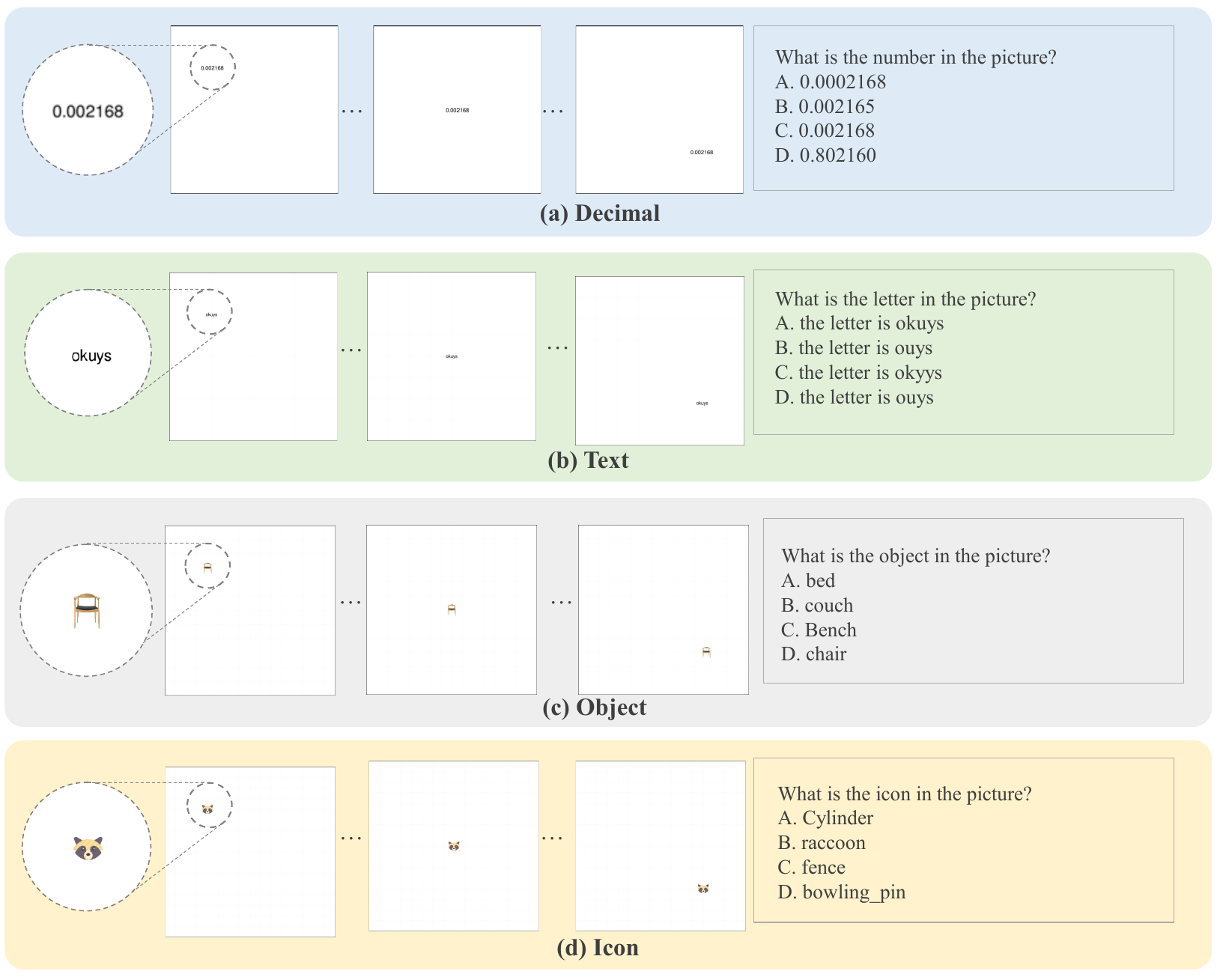}
\caption{\textbf{Examples of our proposed EntityGrid-QA Benchmark.}
}
\label{fig:benchmarkexample}
\end{center}
\end{figure*}

\paragraph{Samples from EntityGrid-QA Benchmark.} We illustrate three examples from our proposed EntityGrid-QA benchmark in \cref{fig:benchmarkexample}. These four samples visualize examples of the four tasks in the benchmark we proposed. For each task, we write or paste the digital number or object directly onto each position of an empty image, and ask questions to the models.

\begin{figure*}[!ht]
\begin{center}
\includegraphics[width=0.75\linewidth]{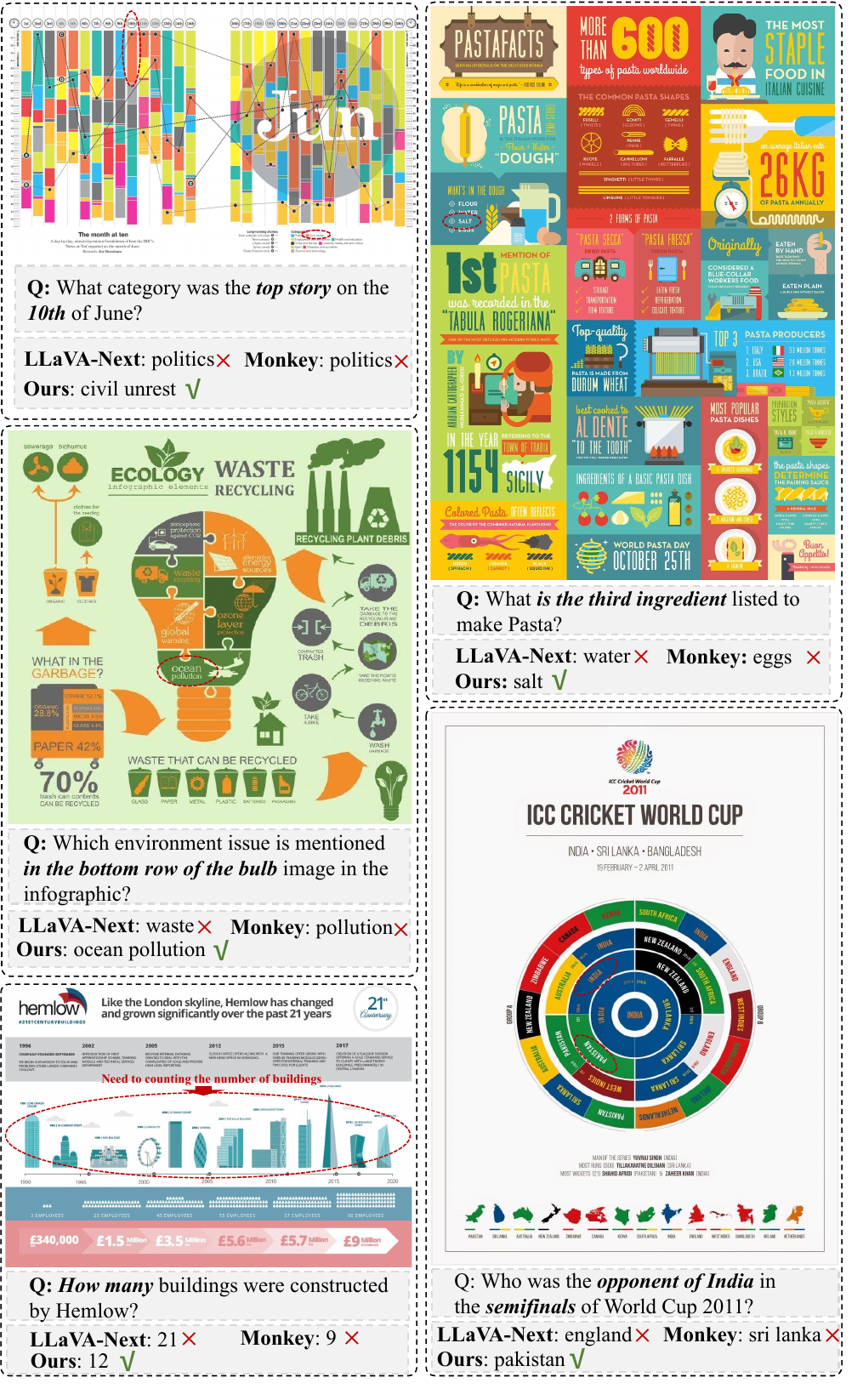}
\caption{\textbf{Qualitative results from InfoVQA~\citep{mathew2022infographicvqa}.}
}
\label{fig:appendix_visualization_infovqa}
\end{center}
\end{figure*}

\begin{figure*}[!ht]
\begin{center}
\includegraphics[width=0.75\linewidth]{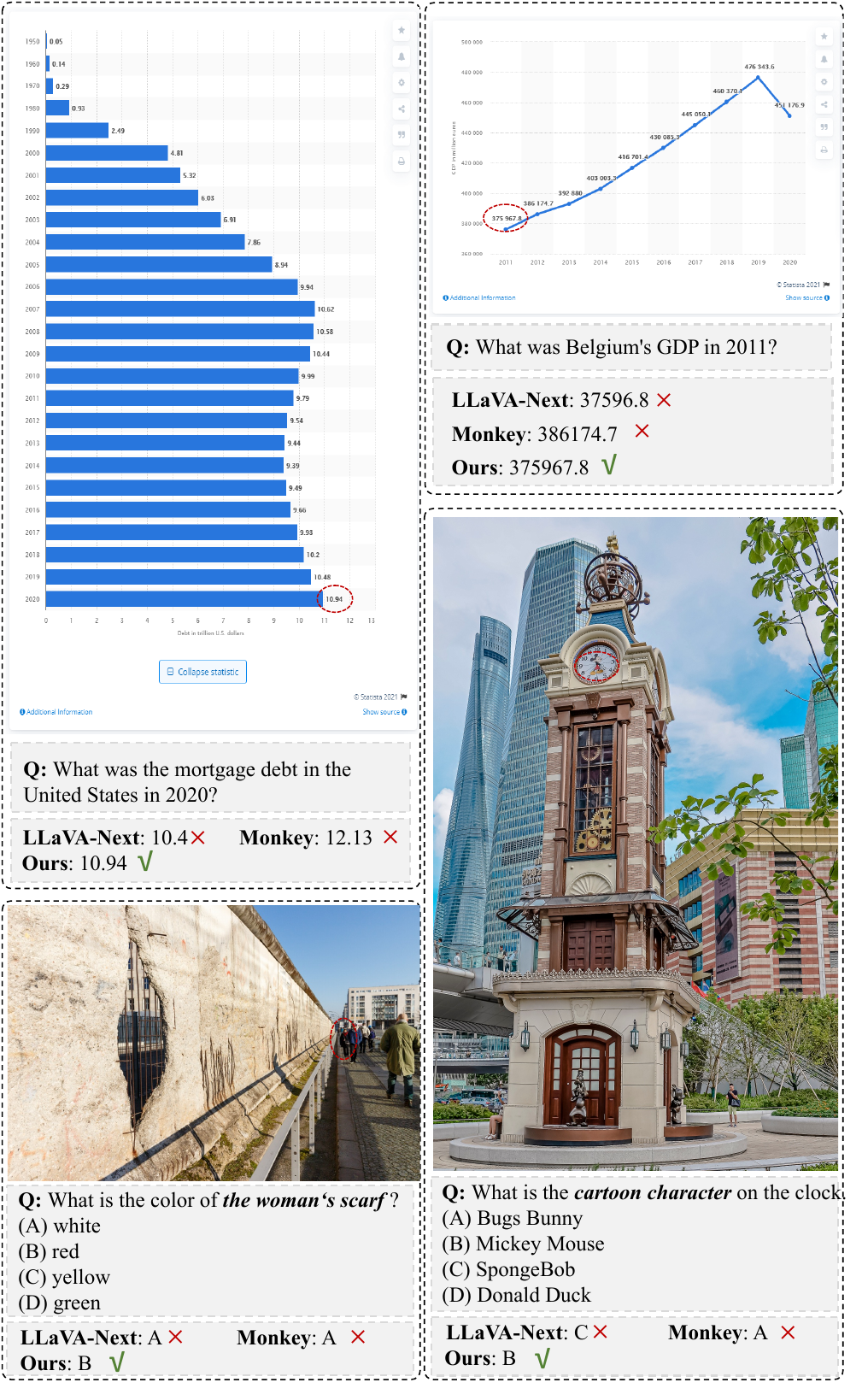}
\caption{\textbf{Qualitative results from ChartQA~\citep{masry2022chartqa} and Vstar Benchmark~\citep{wu2023vstar}.}
We use the red circle to highlight the answer target in the image.
}
\label{fig:appendix_visualization_chartqa_vstar}
\end{center}
\end{figure*}

\begin{figure*}[!ht]
\begin{center}
\includegraphics[width=0.75\linewidth]{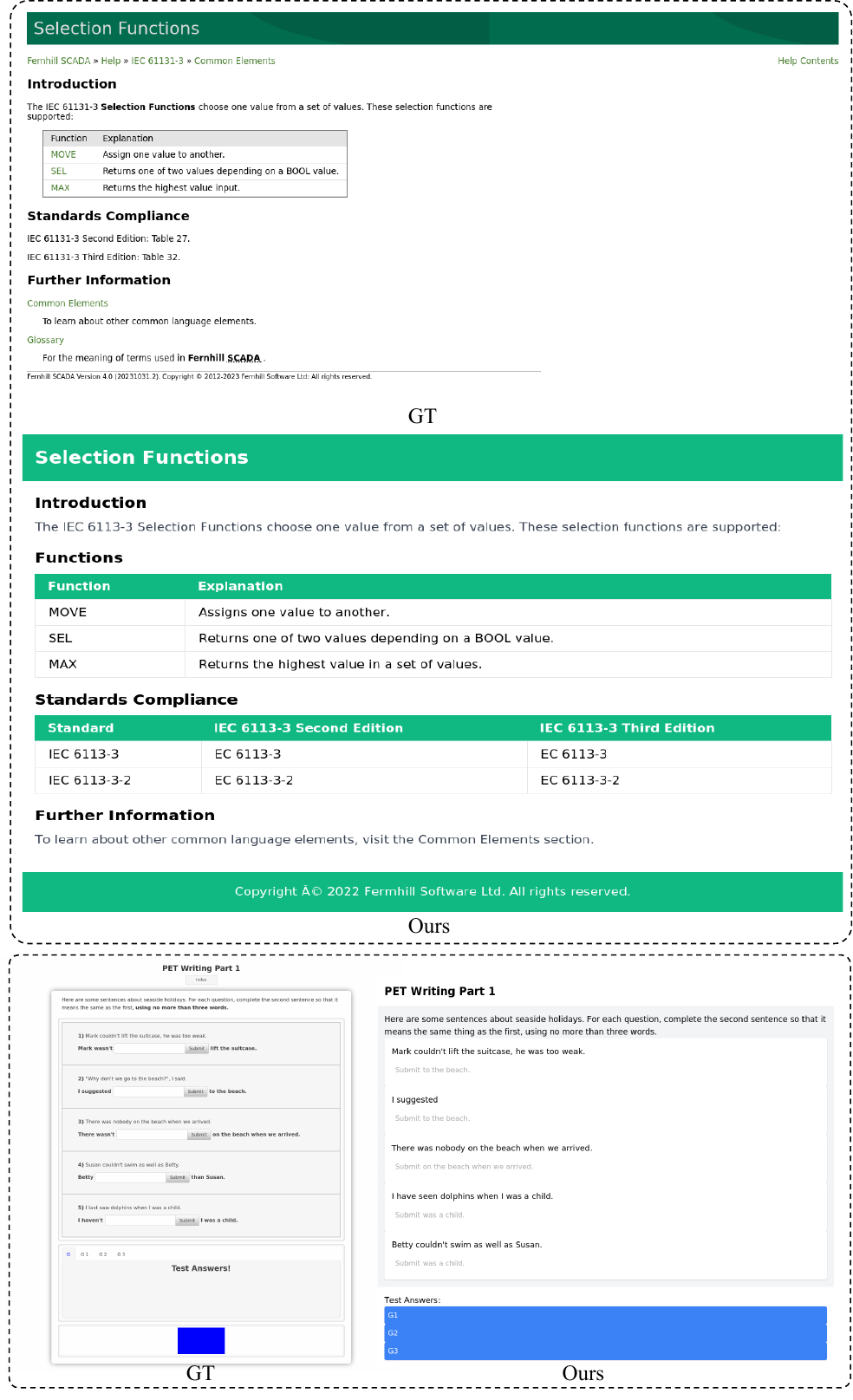}
\caption{\textbf{Qualitative results on Image2HTML task~\citep{si2024design2code}.} We visualize convert the generated html code to website image and compare to the input image.
}
\label{fig:appendix_visualization_design2code}
\end{center}
\end{figure*}

\paragraph{More Qualitative Results.} To further validate the effectiveness of our model, we illustrate the more qualitative results of InfoVQA, ChartQA and V* Benchmark in \cref{fig:appendix_visualization_infovqa} and \cref{fig:appendix_visualization_chartqa_vstar}. Moreover, we give two qualitative examples to present the \method's capability of generating HTML code when given a website image in \cref{fig:appendix_visualization_design2code}.

\section{Broader Impacts}\label{sec:broader_impacts}
The development of \method~advances the field of vision-language models and has broad implications for various applications, including document analysis, medical imaging and remote sensing. However, alongside these potential benefits, there are considerable concerns.

\method, not having undergone rigorous safety training, might generate harmful or inappropriate content, leading to legal and ethical issues. Furthermore, its enhanced ability to process high-resolution inputs could be misused for creating misleading news, contributing to disinformation. These potential negative impacts highlight the need for careful management and ethical guidelines in the deployment of such technologies.

\end{document}